\documentclass{article}

\usepackage{times}
\usepackage{graphicx} 
\usepackage{subfigure} 

\usepackage{natbib}

\usepackage{algorithm}
\usepackage{algorithmic}

\usepackage[draft]{fixme}

\usepackage{amsthm}
\newtheorem{thm}{Theorem}
\newtheorem{lem}{Lemma}
\newtheorem{cor}{Corollary}

\usepackage{hyperref}

\usepackage[accepted]{icml2017}

\icmltitlerunning{Bayesian Models of Data Streams with HPPs}

\usepackage{hl_short}
\usepackage{amssymb}
\usepackage{tikz}
\usetikzlibrary{trees}
\usetikzlibrary{shadows}
\usetikzlibrary{backgrounds}
\usetikzlibrary{shapes,arrows,automata}
\usetikzlibrary{positioning,intersections,fit,calc}
\usetikzlibrary{bayesnet}

\usepackage{enumitem}

\usepackage{bm}
\usepackage{amsmath}
\nc{\Pa}{{\rm pa}}
\nc{\Ch}{{\rm ch}}
\nc{\Co}{{\rm co}}
\nc{\MB}{{\rm mb}}
\nc{\dataset}{\calD}

\renewcommand{\KL}{\emph{KL}}

\nc{\lb}{{\cal L}}
\nc{\llb}{{\cal \hat{L}}}
\nc{\Exp}{\mathbb{E}}
\newcommand\notype[1]{\unskip}
\newcommand{\testData}{\ensuremath{\tilde{\bmx}_t}\xspace}

\begin{document} 

\twocolumn[
\icmltitle{Bayesian Models of Data Streams with Hierarchical Power Priors}




\begin{icmlauthorlist}
\icmlauthor{Andr\'{e}s Masegosa}{ual,ntnu}
\icmlauthor{Thomas D. Nielsen}{aau}
\icmlauthor{Helge Langseth}{ntnu}
\icmlauthor{Dar\'{\i}o Ramos-L\'{o}pez}{ual}
\icmlauthor{Antonio Salmer\'{o}n}{ual}
\icmlauthor{Anders L. Madsen}{aau,hug}
\end{icmlauthorlist}

\icmlaffiliation{ual}{Department of Mathematics, Unversity of Almer\'{\i}a, Almer\'{\i}a, Spain}
\icmlaffiliation{aau}{Department of Computer Science, Aalborg University, Aalborg, Denmark}
\icmlaffiliation{ntnu}{Department of Computer and Information Science, Norwegian University of Science and Technology, 
Trondheim, Norway}
\icmlaffiliation{hug}{Hugin Expert A/S, Aalborg, Denmark}
\icmlcorrespondingauthor{Andr\'{e}s Masegosa}{andresmasegosa@ual.es}

\icmlkeywords{data streams, Bayesian model, adaptivity, power priors, concept drift}

\vskip 0.3in
]



\printAffiliationsAndNotice{}  

\begin{abstract} 
Making inferences from data streams is a pervasive problem in many modern data analysis applications. But it requires to address the problem of continuous model updating, and adapt to changes or drifts in the underlying data generating distribution. In this paper, 
we approach these problems from a Bayesian perspective covering general conjugate exponential models.
Our proposal 
makes use of non-conjugate hierarchical priors to explicitly model temporal changes of the model parameters. We also derive a novel variational inference scheme which overcomes the use of non-conjugate priors while maintaining the computational efficiency of variational methods over conjugate models. The approach is validated on three real data sets over three latent variable models. 
\end{abstract} 

\section{Introduction}\label{sec:Introduction}

Flexible and computationally efficient models for streaming data are required in many machine learning applications, 
and in this paper we propose a new class of models for these situations. 
Specifically, we are interested in models suitable for domains that exhibit changes in the underlying generative process \cite{GamaZliobaiteBifetPechenienizkiyBouchachia14}.
We  envision a situation, where one receives batches of data at discrete points in time. 
As each new batch arrives, we want to glean information from the new data, while also retaining relevant information from the historical observations. 

Our modelling is inspired by previous works on \textit{Bayesian recursive estimation} \cite{OzkanSmidlSahaLundquistGustafsson13,karny2014approximate},  \textit{power priors} \cite{ibrahim2000power} and exponential 
forgetting approaches \cite{honkela2003line}. 
However, all of these methods were developed for  slowly changing processes, where the rate of change anticipated by the model is controlled by a 
quantity that must be set manually. 
Our solution, on the other hand, can accommodate both gradual and abrupt concept drift by continuously assessing the similarity between new and historic data 
using a fully Bayesian paradigm.  

Building Bayesian models for  data streams raises computational problems, as data may arrive with high velocity and is unbounded in size. 
We therefore develop an approximate variational inference technique based on a novel lower-bound of the data likelihood function.  
The appropriateness of the approach is investigated through experiments using both synthetic and real-life data, giving encouraging results. 
The proposed methods are released as part of an open-source toolbox for scalable probabilistic machine learning (\url{http://www.amidsttoolbox.com}) \citep{masegosa2017amidst,masegosa2016probabilistic,cabanas2016financial}.

\vekk{
The remainder of this paper is organized as follows: 
\secref{Preliminaries} presents preliminaries for the developments, followed by a  brief discussion of related work in \secref{RelatedWork}.  
The proposed model is detailed in \secref{HierarchicalPowerPriors}, before we report on the results of a number of experiments in \secref{Experiments}. 
The paper is concluded with some pointers to future research in \secref{ConclusionsAndFutureWork}.\fixme{Thomas always claims that this paragraph is not needed...}
}

\section{Preliminaries}\label{sec:Preliminaries}

\begin{figure}
\begin{center}
\begin{tabular}{cc}
\scalebox{0.95}{
\begin{tikzpicture}
  \node[latent]  (beta)   {$\boldsymbol{\beta}$}; %
  
  \node[const, left= .5  of beta]  (alpha)   {$\bmalpha$}; %
  \edge{alpha}{beta};
  
  \node[obs, below right=1 of beta]          (X)   {$\bmx_{i}$}; %
  \edge{beta}{X};
 
  \node[latent, below left= 1 of beta]          (Z)   {$\bmz_{i}$}; %
  \edge{Z}{X};
  \edge{beta}{Z};
  
  \plate {Observations} { %
    (X)(Z)
  } {\tiny $i={1,\ldots, N}$} ;


\end{tikzpicture}
}
&
\scalebox{0.85}{
\begin{tikzpicture}
  \nc{\nsize}{0.9cm}
  
  \node[latent, minimum size=\nsize]  (beta)   {$\boldsymbol{\beta}_t$}; %
  
  \node[latent, left=.5 of beta, minimum size=\nsize] (beta-1)  {$\boldsymbol{\beta}_{t-1}$}; %
  \edge{beta-1}{beta};
  
  \node[latent, above=.5 of beta, minimum size=\nsize] (rho)  {$\boldsymbol{\rho_t}$}; %
  \edge{rho}{beta};

  \node[const, left=.5 of rho] (gamma)  {$\gamma$}; %
  \edge{gamma}{rho};

  \node[obs, below right=1 of beta, minimum size=\nsize]          (X)   {$\bmx_{i,t}$}; %
  \edge{beta}{X};
 
  \node[latent, below left= 1 of beta, minimum size=\nsize]          (Z)   {$\bmz_{i,t}$}; %
  \edge{Z}{X};
  \edge{beta}{Z};
  
  \plate {Observations} { %
    (X)(Z)
  } {\tiny $i={1,\ldots, N}$} ;


\end{tikzpicture}
}\\
(a) & (b)
\end{tabular}
\end{center}
\vspace{-3mm}
\caption{\label{fig:plateModel} Left figure displays the core of the probabilistic model examined in this paper. Right figure includes a   temporal evolution model for $\bmbeta_t$ as described in \secref{HierarchicalPowerPriors}.}
\end{figure}
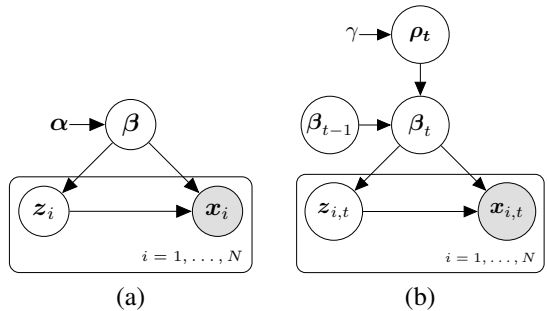

In this paper we focus on conjugate exponential Bayesian network models for performing Bayesian learning on
streaming data. 
To simplify the presentation, we shall initially focus on the model structure shown in \figref{plateModel}~(a).
This model includes the observed data $\bmx = \bmx_{i=1:N}$, global hidden
variables (or parameters)
$\boldsymbol{\beta} = \bmbeta_{1:M}$, a set of local hidden variables $\bmz =
\bmz_{1:N}$, and a vector of fixed (hyper) parameters denoted by $\bmalpha$.  
%
Notice how the dynamics of the process is not included in the model of \figref{plateModel}~(a); the model will be set in the context of data streams in \secref{HierarchicalPowerPriors}, 
where we extend it to incorporate explicit dynamics over the (global) parameters to capture concept drift.


With the conditional distributions in the model belonging to the exponential family, we have that all
distributions are of the following form 
\begin{equation}\nonumber
\begin{split}
& \ln p(Y|\Pa(Y)) = \\
&\phantom{=} \ln h_Y  + \bmeta_{Y}(\Pa(Y))^T \bmt_Y(Y) - a_{Y}(\bmeta_Y(\Pa(Y))),
\end{split}
\end{equation}
\noindent where $\Pa(Y)$ denotes the parents of $Y$ in the directed acyclic graph of the induced Bayesian network
model. The scalar functions $h_Y$ and $a_Y(\cdot)$ are the base measure and the log-normalizer,
respectively; the vector functions $\bmeta_Y(\cdot)$ and $\bmt_Y(\cdot)$ are the \textit{natural parameters} and
the \textit{sufficient statistics} vectors, respectively. The subscript $Y$ means that the associated
functional forms may be different for the different factors of the model, but we may remove the subscript when
clear from the context. 
By also requiring that the distributions are conjugate, we have that the posterior distribution for each
variable in the model has the same functional form as its prior distribution. Consequently, learning (i.e.\
conditioning the model on observations) only changes the values of the parameters of the model, and not the
functional form of the distributions. \vekk{This can be achieved by expressing the functional form of $p(Y|\Pa(Y))$
in terms of the sufficient statistics $\bmt_Z(Z)$ of any of the parents $Z\in \Pa(Y)$ of $Y$:
\begin{equation}\nonumber
\begin{split}
&\ln p(Y|\Pa(Y)) =\\
&\phantom{} \ln h_Z  + \bmeta_{YZ}(Y,\Co_Z(Y))^T \bmt_Z(Z) - a_{Z}(\bmeta_{YZ}(X,\Co_Z(Y))),
\end{split}
\end{equation}
\noindent where $\Co_Z(Y)$ denotes the coparents of $Z$ with respect to $Y$, i.e. $\Co_Z(Y)=\Pa(Y)\setminus \{Z\}$.
}

Variational inference is a deterministic
technique for finding tractable posterior distributions, denoted by $q$, which approximates the Bayesian
posterior,  $p(\bmbeta,\bmz|\bmx)$, that is often intractable to compute. More specifically, by letting ${\cal Q}$
be a set of possible approximations of this posterior, variational inference solves the following optimization problem for any model in the conjugate exponential family:
\begin{equation} \label{eq:VI}
\min_{q\left(\bmbeta,\bmz\right)\in {\cal Q}} \KL(q(\bmbeta,\bmz)|p(\bmbeta,\bmz|\bmx)),
\end{equation}
\noindent where $\KL$ denotes the Kullback-Leibler divergence between two probability distributions.

In the \textit{mean field variational} approach the approximation family ${\cal Q}$ is assumed to fully factorize. 
Extending  the notation of  \citet{HoffmanBleiWangPaisley13}, we have that
\[
q(\bmbeta,\bmz\given \bmlambda,\bmphi) = \prod_{k=1}^M q(\beta_{k}\given \lambda_{k})\prod_{i=1}^N\prod_{j=1}^J q(z_{i,j}\given \phi_{i,j}),
\] 
where $J$ is the number of local hidden variables, which is assumed fixed for all $i=1,\ldots, N$. 
The parameterizations of the variational distributions are made explicit, in that $\bmlambda$ parameterize the variational distribution of $\bmbeta$, while $\bmphi$ has the same role for the variational distribution of $\bmz$.

To solve the minimization problem in Equation (\ref{eq:VI}), the variational approach exploits  the  transformation
\begin{equation}
  \label{eq:likelihood_decomposition}
\ln P(\bmx) = {\cal L}(\bmlambda,\bmphi\given \bmx, \bmalpha_u) + \KL(q(\bmbeta,\bmz\given \bmlambda,\bmphi)|p(\bmbeta,\bmz|\bmx)),  
\end{equation}
\noindent where ${\cal L}(\cdot|\cdot)$ is a \emph{lower bound} of $\ln P(\bmx)$ since $\KL$ is  non-negative. 
$\bmx$ and $\bmalpha_u$ are introduced in \calL's notation to make explicit the function's dependency on  $\bmx$, the data sample, and $\bmalpha_u$, the natural parameters of the prior over $\bmbeta$. 
As  $\ln P(\bmx)$ is constant, minimizing the $\KL$ term is equivalent to maximizing the lower bound. 
Variational methods maximize this lower bound by applying a coordinate  ascent that iteratively updates the individual variational distributions while holding the others fixed~\citep{WinnBishop05}. 
The key advantage of having a conjugate exponential model is that the gradients of the $\lb$ function can be always computed in closed form ~\citep{WinnBishop05}.

\section{Related Work}\label{sec:RelatedWork}

Bayesian inference on streaming data has been widely studied \cite{ahmed2011online,doucet2000sequential,yao2009efficient}. 
In the context of variational inference, there are two main approaches. \citet{ghahramani2000online,BroderickBoydWibisonoWilsonJordan13} propose recursive Bayesian updating of the variational approximation. 
The streaming variational Bayes (SVB) algorithm \cite{BroderickBoydWibisonoWilsonJordan13}  is the most known approach of this category. 
Alternatively, one could cast the inference problem as a stochastic optimization problem. 
Stochastic variational inference (SVI) \cite{HoffmanBleiWangPaisley13} and the closely related population variational Bayes (PVB) \cite{McInerneyRanganathBlei15} 
are prominent examples from this group. 
 SVI assumes the existence of a fixed data set observed in a sequential manner, and in particular that this data set has a known finite size.
This is unrealistic  when modeling data streams. 
PVB addresses this problem by using the frequentist notion of a population distribution, $\mathbf{F}$,  which is assumed to generate the data stream by repeatedly sampling 
$M$ data points at the time.  
$M$ parameterizes the size of the population, and helps control the variance of the population posterior. 
Unfortunately, $M$ must be specified by the user. No clear rule exists regarding how to set it, and
 \citet{McInerneyRanganathBlei15} show that its optimal value may differ from one data stream to another. 

The problem of Bayesian modeling of non-stationary data streams (i.e.,  with concept drift \cite{GamaZliobaiteBifetPechenienizkiyBouchachia14}) is not addressed by SVB, as it assumes data exchangeability. 
An online variational inference method, which exponentially forgets the variational parameters associated with old data, was proposed by \citet{honkela2003line}. 
The so-called \textit{power prior} approach \cite{ibrahim2000power} is also based on an exponential forgetting mechanisms, and has nice theoretical properties \cite{ibrahim2003optimality}. 
Nevertheless, both approaches rely on a hyper-parameter determining forgetting, which has to be set manually. 
PVB can also adapt to concept drift, because the variance of the variational posterior never decreases below a given threshold indirectly controlled by ${M}$, but again, the hyper-parameter has to be set manually. 

A time series based modeling approach for concept drift using implicit transition models was pursued by \citet{OzkanSmidlSahaLundquistGustafsson13,karny2014approximate}.  Unfortunately, the implicit transition model depends on a hyper-parameter determining the forgetting-factor, which has to be manually set. 
In this paper we build on this approach, adapt it to variational settings, and place a hierarchical prior on its forgetting parameter. 
This greatly improves the flexibility and accuracy of the resulting model when making inferences over drifting data streams.

\vekk{
A number of alternatives pursuing Bayesian modeling of streaming data can be found in the literature.
Streaming Variational Bayes (SVB) \cite{BroderickBoydWibisonoWilsonJordan13} is based on a straightforward 
adaptation of Bayes theorem,
\begin{equation}
\label{eq:SVB}
p(\bmbeta,\bmz|\bmd_1,\ldots,\bmd_N) \propto p(\bmbeta,\bmz) \prod_{i=1}^N p(\bmbeta,\bmz_k|\bmd_i)  .
\end{equation}

{
    \def\OldComma{,}
    \catcode`\,=13
    \def,{%
      \ifmmode%
        \OldComma\discretionary{}{}{}%
      \else%
        \OldComma%
      \fi%
    }%
The above expression shows that the Bayesian posterior, $p(\bmbeta,\bmz|\bmd_1,\ldots,\bmd_N)$, can be built by independently combining the intermediate posteriors, $p(\bmbeta,\bmz_k|\bmd_k)$. The SVB approach is based on building local variational approximations $q_{k}(\bmbeta,\bmz_k)$ to the exact (intractable) intermediate posteriors $p(\bmbeta,\bmz_k|\bmd_k)$ for each data point and, then, combining them following Equation \ref{eq:SVB} to obtain a global approximate posterior, $q(\bmbeta,\bmz)$.  When the model is conjugate exponential, this combination rule reduces to updating the natural parameters as
}
\begin{equation}
\label{eq:SVB2}
\bmeta^q = \bmeta^p_{prior} + \sum_{k=1}^m (\bmeta^q_k - \bmeta^p_{prior}) .
\end{equation}
But, unless the local variational approximations provide an exact approximation, i.e., $\KL(q_k,p)=0$,  the SVB approach does not yield an optimal variational solution. Furthermore, it does not even guarantee convergence to a stationary point of the lower bound.

A different approach is adopted by the Population Variational Bayes (PVB) method \cite{McInerneyRanganathBlei15}.
PVB assumes the data stream to be composed by minibatches of size $ESS$ (equivalent sample size). The data-generating distribution
is denoted by $F$. Rather than maximizing the
lower bound, PVB targets the function
\begin{multline}
\label{eq:flowerbound}
{\cal L}(\lambda,\phi;F) = \mathbb{E}_{F}\left[  \mathbb{E}_q \left[  \log p(\bmbeta) - \log q(\bmbeta|\bmlambda) + \right. \right. \\
\sum_{i=1}^{ESS} \log p(\bmx_i,\bmz_i|\bmbeta) - \log q(\bmz_i)  ]] .
\end{multline}
The process starts by randomly initializing $\bmlambda$ and then finding the $\bmphi$ values that maximize Equation~\eqref{eq:flowerbound}.
The update rule for $\bmlambda$ is
\begin{multline}
\label{eq:updatepvb}
\bmlambda^{(k)} = \bmlambda^{(k-1)} + \\
\nu \frac{ESS}{B}\left( \bmalpha - \bmlambda^{(k-1)} + \mathbb{E}_{F} \left[\sum_{i=1}^{ESS} \mathbb{E}_{\mathbf{\bmphi^{(k-1)}}}[t(\bmx_i,\bmz_i)]\right] \right), 
\end{multline}
where $\nu$ is the learning rate and $B$ is the number of items used to estimate the expectations, see  \cite{McInerneyRanganathBlei15}.

The $ESS$ parameter helps control the variance of the population posterior. It is a user-specidfied parameter for which no clear rule exists.
The analysis in \cite{McInerneyRanganathBlei15} shows that the optimal setting of $ESS$ is often different to the
actual number of data points, which is a convenient feature, as the number of data points is not known when analyzing data streams.
 }

\section{Hierarchical Power Priors}\label{sec:HierarchicalPowerPriors}
In this section we extend the model in Figure~\ref{fig:plateModel}~(a) to also account for the dynamics of the
data stream being modeled. We shall here assume that only the parameters $\bmbeta$ in
Figure~\ref{fig:plateModel}~(a) are time-varying, which we
will indicate with the subscript $t$, i.e., $\bmbeta_t$. 
First we briefly describe the approach on which the proposed model is based. Afterwards, we introduce the hierarchical power prior  and detail a variational inference procedure for this model class.

\subsection{Power Priors as Implicit Transition Models}
\label{sec:HPP:power-priors}
In order to extend the model in Figure~\ref{fig:plateModel}~(a) to data streams, we may introduce a transition
model $p(\bmbeta_t\given \bmbeta_{t-1})$ to explicitly model the evolution of the parameters over time,
enabling the estimation of the predictive density at time $t$:
\begin{equation}
  \label{eq:transition_model}
  p(\bmbeta_t\given \bmx_{1:t-1}) = \int p(\bmbeta_t\given \bmbeta_{t-1})p(\bmbeta_{t-1}\given \bmx_{1:t-1})d\bmbeta_{t-1}.  
\end{equation}
However, this approach introduces two  problems. First of all, in non-stationary domains we may not
have a single transition model or the transition model may be unknown. Secondly, if we
seek to position the model within the conjugate exponential family in order to be able to compute the gradients of $\lb$ in closed-form, we need to ensure that the distribution
family for $\bmbeta_t$ is its own conjugate distribution, thereby severely limiting model
expressivity (we can, e.g., not assign a Dirichlet distribution to $\bmbeta_t$).

Rather than explicitly modeling the evolution of the
$\bmbeta_t$ parameters as in Equation~(\ref{eq:transition_model}), we instead follow the approach of
\citet{karny2014approximate} and \citet{OzkanSmidlSahaLundquistGustafsson13} who define the time
evolution model implicitly by constraining the maximum \KL\ divergence over consecutive
parameter distributions. Specifically, by defining 
\begin{equation}
  \label{eq:delta_transition_model}
p_{\delta}(\bmbeta_t\given \bmx_{1:t-1}) = \int \delta(\bmbeta_t-\bmbeta_{t-1})p(\bmbeta_{t-1}\given \bmx_{1:t-1})d\bmbeta_{t-1}
\end{equation}
one can restrict the space of possible distributions $p(\bmbeta_t\given
\bmx_{1:t-1})$, supported by an unknown transition model, by the constraint
\begin{equation}
  \label{eq:KL-constraint}
  \KL(p(\bmbeta_t\given
\bmx_{1:t-1}), p_{\delta}(\bmbeta_t\given \bmx_{1:t-1})) \leq \kappa.
\end{equation}
\citet{karny2014approximate} and \citet{OzkanSmidlSahaLundquistGustafsson13} seek to approximate $p(\bmbeta_t\given
\bmx_{1:t-1})$ by the distribution $\hat{p}(\bmbeta_t\given \bmx_{1:t-1})$ having maximum entropy under the
constraint in (\ref{eq:KL-constraint}); for continuous distributions the maximum entropy can be formulated
relative to an uninformative prior density $p_u(\bmbeta_t)$, which  corresponds to the Kullbach-Leibler
divergence between the two distributions. This approach ensures that we will not underestimate the uncertainty
in the parameter distribution and the particular solution being sought takes the form
\begin{equation}
  \label{eq:power-prior}
\hat{p}(\bmbeta_{t}\given \bmx_{1:t-1},\rho_t) \propto p_{\delta}(\bmbeta_t\given \bmx_{1:t-1})^{\rho_t}
p_u(\bmbeta_t)^{(1-\rho_t)},  
\end{equation}
where $0\leq \rho_t \leq 1$ is indirectly defined by (\ref{eq:KL-constraint}) which in turn depends on the user
defined parameter $\kappa$. 

In our streaming data setting we follow \textit{assumed density filtering} \cite{lauritzen1992propagation} and the SVB approach \cite{BroderickBoydWibisonoWilsonJordan13} and employ
the approximation
$
p(\bmbeta_{t-1}\given \bmx_{1:t-1}) \approx q(\bmbeta_{t-1}\given \bmlambda_{t-1}), 
$
where $q(\bmbeta_{t-1}\given \bmlambda_{t-1})$ is the variational distribution calculated in the previous time
step. Using this approximation in (\ref{eq:transition_model}) and (\ref{eq:delta_transition_model}), we can express $p_\delta$ in
terms of $\bmlambda_{t-1}$ in which case (\ref{eq:power-prior}) becomes 
\begin{equation}
\label{eq:PowerPrior}
\hat{p}(\bmbeta_{t}\given \bmlambda_{t-1},\rho_t) \propto p_{\delta}(\bmbeta_t\given \bmlambda_{t-1})^{\rho_t} p_u(\bmbeta_t)^{(1-\rho_t)},  
\end{equation}
which we use as the prior density for time step $t$. Now, if $p_u(\bmbeta_t)$ belong to the same family as
$q(\bmbeta_{t-1}\given \bmlambda_{t-1})$, then $\hat{p}(\bmbeta_{t}\given
\bmlambda_{t-1},\rho_t)$ will stay within the same family and have natural parameters $\rho_t \bmlambda_{t-1} +
(1-\rho_t)\bmalpha_u$, where $\bmalpha_u$ are the natural parameters of $p_u(\bmbeta_t)$. Thus, under this
approach, the transitioned posterior remains within the same exponential family, so we can enjoy the full flexibility of the conjugate exponential family (i.e. computing gradients of the $\lb$ function in closed form), an option that would not be available if one were to explicitly specify a transition model as in Equation~(\ref{eq:transition_model}). 

So, at each time step, we simply have to solve the following variational problem, where only the prior changes with respect to the original SVB approach,
\[
\arg\max_{\bmlambda_t,\bmphi_t} \lb(\bmlambda_t,\bmphi_t|\bmx_t,\rho_t\bmlambda_{t-1} + (1-\rho_t)\bmalpha_u).
\]
As stated in the following lemma, this approach coincides with the so-called \emph{power priors} approach
\cite{ibrahim2000power}, a term that we will also adopt in the following. 
\begin{lem}
\label{lem:pp}
The Bayesian updating scheme described by Figure~\ref{fig:plateModel}~(b) and
Equation~\ref{eq:power-prior}, but with $\rho_t$ fixed to a constant value, is equivalent to the recursive application of the Bayesian updating scheme of power priors \cite{ibrahim2000power}. This scheme is expressed as follows:
\[
p(\bmbeta|\bmx_1,\bmx_0,\rho) \propto p(\bmx_1\given \bmbeta) p(\bmx_0\given \bmbeta)^\rho p(\bmbeta),
\]
\noindent where $\bmx_0$ and $\bmx_1$ is the observation at time $0$ (historical observation) and time $1$
(current observation), respectively.
\end{lem}
\begin{proof}[Proof sketch] Translate the recursive Bayesian updating approach of power priors into an equivalent two time slice model, where   $\bmbeta_0$ is given a prior distribution $p$ and $p(\bmbeta_1\given \bmbeta_0)$ is a Dirac delta
  function. The distribution $p(\bmbeta_1\given \bmx_0,\bmx_1,\rho)$ in this model is equivalent to
  $p(\bmbeta|\bmx_1,\bmx_0,\rho)$, which, in turn, is equivalent (up to proportionality) to $p(\bmx_1\given \bmbeta_1)\hat{p}(\bmbeta_{1}\given \bmx_0,\rho_t)$. Note that the last $\hat{p}$ term can alternatively be expressed as $\hat{p}(\bmbeta_{1}\given \bmx_0,\rho_t)\propto p_\delta (\beta_1\given \bmx_0)^\rho p(\beta_1)^{1-\rho}\propto p_\delta(\bmx_0|\beta_1)^\rho p(\beta_1)$.
\end{proof}

The perspective provided by Lemma~\ref{lem:pp} introduces a well known result of power priors, which is also applicable in the current context
(see the discussion after Theorem 1 in 
\cite{ibrahim2003optimality}): ``\textit{the power prior is an optimal prior to use and
  in fact minimizes the convex combination of KL divergences between two extremes: one in
  which no historical data is used and the other in which the historical data and current data
  are given equal weight.}'' As noted in \cite{olesen1992ahugin,OzkanSmidlSahaLundquistGustafsson13}, this
schema works as a moving window with exponential forgetting of past data, where the effective number of samples or, more technically, the so-called \textit{equivalent sample size} of the posterior \cite{heckerman1995learning},  converges to,
\begin{equation}
\label{eq:ESSLimit}
\lim_{t\rightarrow\infty} ESS_t = \frac{|\bmx_t|}{1-\rho}
\end{equation}
\noindent if the size of the data batches is constant\footnote{For instance, the ESS of a Beta distribution is equal to the sum of the components of $\bmlambda_t$ and, in turn, equal to the number of data samples seen so far plus the prior's pseudo-samples.}. 

For the experimental results reported in Section~\ref{sec:Experiments} we shall refer to the method outlined above as SVB with power priors (SVB-PP).    



%
%
%
%
%
%

\subsection{The Hierarchical Power Prior Model}
In the approach taken by \citet{OzkanSmidlSahaLundquistGustafsson13} (and, by extension, SVB-PP), the forgetting factor $\rho_t$ is
user-defined. In this paper, we instead pursue a (hierarchical) Bayesian approach and
introduce a prior distribution over $\rho_t$ allowing the distribution over $\rho_t$ (and thereby the forgetting
mechanism) to adapt to the data stream. 

As we shall see below, in order to support a variational updating
scheme we need to restrict the prior distribution over $\rho_t$, effectively limiting the choice of prior distribution to either an
exponential distribution or a normal distribution with fixed variance, both of which should be truncated to the
interval $[0,1]$. Unless explicitly stated otherwise, we shall for now assume a truncated exponential distribution with
natural parameter $\gamma$ as prior distribution over $\rho_t$: 
\begin{equation}
\label{eq:truncatedExp}
p(\rho_t\given \gamma) = \frac{\gamma \exp(-\gamma\rho_t)}{1-\exp(-\gamma)}.
\end{equation}

The resulting model can be illustrated
as in Figure~\ref{fig:plateModel}~(b). We shall refer to models of this type as \emph{hierarchical
power prior} (HPP) models.



\vekk{
\begin{figure}
\begin{center}
\scalebox{1}{
\begin{tikzpicture}
  \nc{\nsize}{0.9cm}
  
  \node[latent, minimum size=\nsize]  (beta)   {$\boldsymbol{\beta}_t$}; %
  
  \node[latent, left=.5 of beta, minimum size=\nsize] (beta-1)  {$\boldsymbol{\beta}_{t-1}$}; %
  \edge[dashed]{beta-1}{beta};
  
  \node[latent, above=.5 of beta, minimum size=\nsize] (rho)  {$\boldsymbol{\rho_t}$}; %
  \edge{rho}{beta};

  \node[const, left=.5 of rho] (gamma)  {$\gamma$}; %
  \edge{gamma}{rho};

  \node[obs, below right=1 of beta, minimum size=\nsize]          (X)   {$\bmx_{i,t}$}; %
  \edge{beta}{X};
 
  \node[latent, below left= 1 of beta, minimum size=\nsize]          (Z)   {$\bmz_{i,t}$}; %
  \edge{Z}{X};
  \edge{beta}{Z};
  
  \plate {Observations} { %
    (X)(Z)
  } {\tiny $i={1,\ldots, N}$} ;


\end{tikzpicture}
}
\end{center}
\caption{\label{fig:temporalPlateModel} The model in Figure~\ref{fig:plateModel} extended to include a
  parameter evolution model for $\bmbeta_t$. The parameter model is based on a implicit transition model
  \cite{OzkanSmidlSahaLundquistGustafsson13} with the dashed arc indicating that $\bmbeta_t$ is being defined
  by the natural parameters of $\bmbeta_{t-1}$. }
\end{figure}
}
\subsection{Variational Updating}

For updating the model distributions we pursue a variational approach, where we seek to maximize the evidence lower bound
$\lb$ in Equation~(\ref{eq:likelihood_decomposition}) for time step $t$. However, since the
model in Figure~\ref{fig:plateModel}~(b) does not define a conjugate exponential distribution due to the
introduction of $p(\rho_t)$ we cannot maximize $\lb$ directly. Instead we will derive a (double) lower bound $\llb$
($\llb\leq \lb$) and use this lower bound as a proxy for the updating rules for the variational
posteriors. 

First of all, by instantiating the lower bound $\lb_{HPP}(\bmlambda_t,\bmphi_t,\omega_t|\bmx_t,\bmlambda_{t-1}) $ in
Equation~(\ref{eq:likelihood_decomposition}) for the HPP model we obtain
\begin{equation}
  \label{eq:lower_bound}
\begin{split}
\lb&_{HPP}(\bmlambda_t,\bmphi_t,\omega_t|\bmx_t,\bmlambda_{t-1}) = \Exp_q [\ln p(\bmx_t,\bmZ_t\given \bmbeta_t)] \\
&+ \Exp_q[\ln
\hat{p}(\bmbeta_t\given \bmlambda_{t-1},\rho_t)]  \\
&+ \Exp_q[p(\rho_t\given \gamma)]  - \Exp_q[\ln q(\bmZ_t\given
\bmphi_t)] \\
&- \Exp_q[q(\bmbeta_t\given \bmlambda_t)] - \Exp_q[q(\rho_t\given \omega_t)],  
\end{split}  
\end{equation}
where $\omega_t$ is the variational parameter for the variational distribution for $\rho_t$; as we shall see
later, $\omega_t$ is a scalar and is therefore not shown in boldface. For ease of presentation we shall
sometimes drop from $\lb_{HPP}(\bmlambda_t,\bmphi_t,\omega_t|\bmx_t,\bmlambda_{t-1})$ the subscript as well as
the explicit specification of the parameters when these is otherwise clear from the context.

We now define $\llb_{HPP}(\bmlambda_t,\bmphi_t,\omega_t|\bmx_t,\bmlambda_{t-1})$ as
\begin{align}
  \label{eq:lower_lower_bound}
\llb&_{HPP}(\bmlambda_t,\bmphi_t,\omega_t|\bmx_t,\bmlambda_{t-1}) = \Exp_q [\ln p(\bmx_t,\bmZ_t\given \bmbeta_t)] \nonumber\\
&+ \Exp_q[\rho_t]\Exp_q[\ln p_\delta(\bmbeta_t|\bmlambda_{t-1})] + (1-\Exp_q[\rho_t])\Exp_q[\ln p_u(\bmbeta_t)] \nonumber\\
& + \Exp_q[p(\rho_t\given \gamma)] - \Exp_q[\ln q(\bmZ_t\given \bmphi_t)] \nonumber \\
&- \Exp_q[q(\bmbeta_t\given \bmlambda_t)] - \Exp_q[q(\rho_t\given \omega_t)],
\end{align}

which provide a lower bound for $\lb$.

\begin{thm}
$\llb_{HPP}$ gives a lower bound for $\lb_{HPP}$:
\[
\llb_{HPP}(\bmlambda_t,\bmphi_t,\omega_t|\bmx_t,\bmlambda_{t-1})\leq\lb_{HPP}(\bmlambda_t,\bmphi_t,\omega_t|\bmx_t,\bmlambda_{t-1}).
\]
\end{thm}
\begin{proof}[Proof sketch]
The inequality derives by using Equation~(\ref{eq:bound-difference}) and observing that $a_g(\rho_t\bmlambda_{t-1} + (1-\rho_t)\bmalpha_u) \leq \rho_t a_g(\bmlambda_{t-1}) + (1-\rho_t)a_g(\bmalpha_u)$  because the log-normalizer $a_g$ is always a convex function \cite{wainwright2008graphical}. Full details are given in the supplementary material.
\end{proof}

Rather than seeking to maximize $\lb$ we will instead maximize $\llb$. The gap between the two bounds is
determined only by the log-normalizer of $\hat{p}(\bmbeta_t\given \bmlambda_{t-1},\rho_t)$:
\begin{equation}
  \label{eq:bound-difference}
  \begin{split}
  \llb - \lb = \Exp_q[&\rho_t a_g(\bmlambda_{t-1}) + (1-\rho_t)a_g(\bmalpha_u) \\
&+ a_g(\rho_t\bmlambda_{t-1} + (1-\rho_t)\bmalpha_u)   ]    
  \end{split}
\end{equation}
Thus, maximizing $\llb$ wrt.\ the variational parameters $\bmlambda_t$ and $\bmphi$ also maxmizes $\lb$. By
the same observation, we also have that the (natural) gradients are consistent relative to the two bounds:  
\begin{cor}
  \[
      \hat\nabla_{\bmlambda_t}\lb  =   \hat\nabla_{\bmlambda_t}\llb \qquad
      \hat\nabla_{\bmphi_t}\lb =   \hat\nabla_{\bmphi_t}\llb \, .
  \]
\end{cor}
\begin{proof}
  Follows immediately from Equation~(\ref{eq:bound-difference}) because the difference does not depend of $\bmlambda_t$ and $\bmphi_t$.
\end{proof}
Thus, updating the variational parameters $\bmlambda_t$ and $\bmphi_t$ in HPP models
can be done as for regular conjugate exponential models of the form in Figure~\ref{fig:plateModel}.

In order to update $\omega_t $ we rely on $\llb$, which we can maximize using the natural gradient wrt.\
$\omega_t$ \citep{sato2001online} and which can be calculated in closed form for a restricted distribution
family for $\rho_t$. 
\begin{lem}
\label{lem:GradientOmega}
Assuming that the sufficient statistics function for $\rho_t$ is the identity function, $\bmt(\rho_t)=\rho_t$, then we have 
\begin{equation}
  \label{eq:natural_gradient}
\begin{split}
\hat\nabla _{\omega_t} \llb = & \KL(q(\bmbeta_t\given \bmlambda_t), p_u(\bmbeta_t)) \\
&- \KL(q(\bmbeta_t\given
\bmlambda_t),p_\delta(\bmbeta_t\given \bmlambda_{t-1})) + \gamma - \omega_t
\end{split} 
\end{equation}
\end{lem}
\begin{proof}[Proof sketch] Based on a straightforward algebraic derivation of the gradient using standard properties of the exponential family. Full details are given in the supplementary material. 
\end{proof}

Note that the truncated exponential distribution (see Equation~(\ref{eq:truncatedExp})) satisfies the restriction expressed in Lemma~\ref{lem:GradientOmega}, and also note that the variational posterior $q(\rho_t\given\omega_t)$ will be a truncated exponential density too. 

\begin{figure}[t!]
\vskip -0.2in
\begin{center}
\begin{tabular}{ccc}
\includegraphics[width=9cm,height=5cm,keepaspectratio]{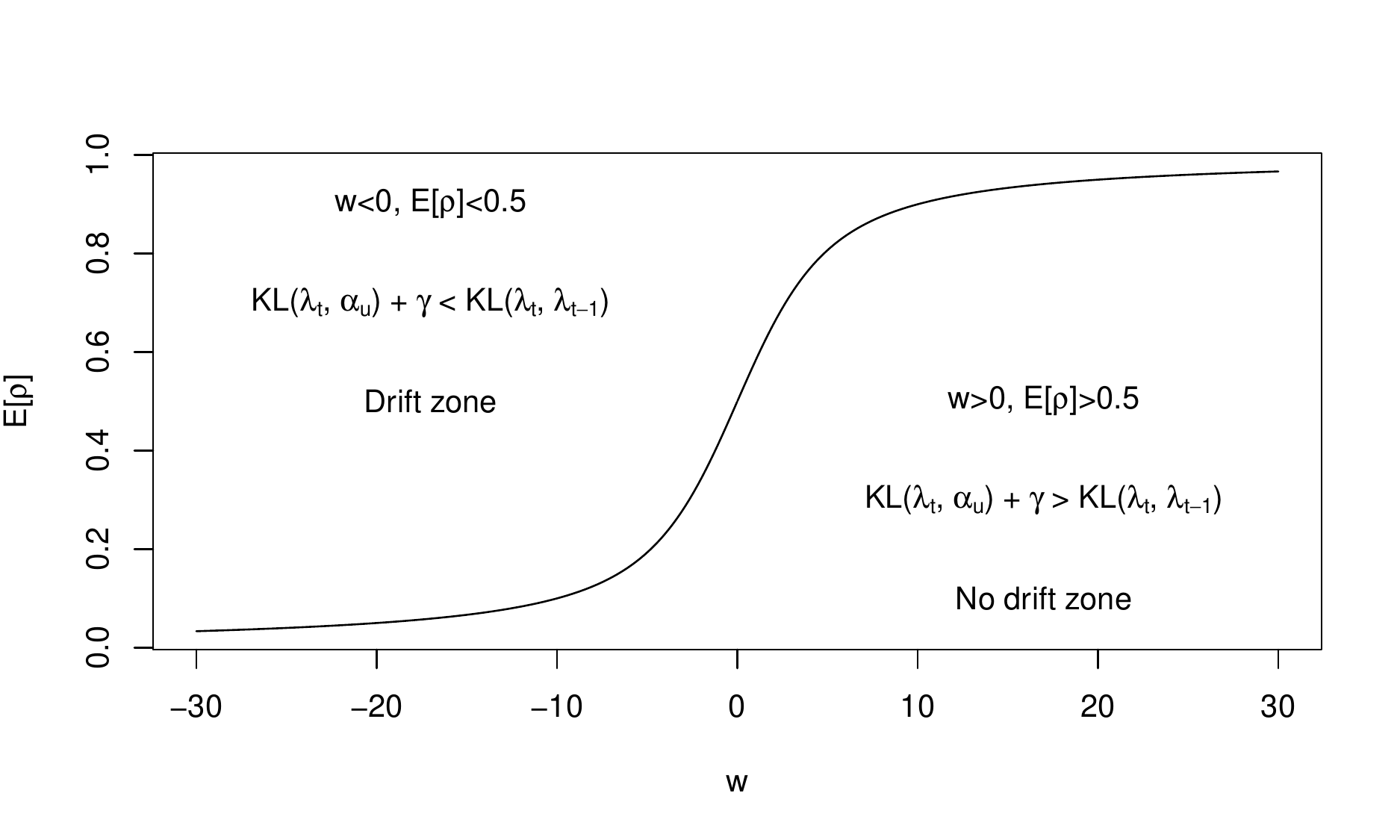}
\end{tabular}
\vskip -0.2in
\caption{Relationship between $\omega_t$ and $E_q[\rho_t]$.} 
\label{fig:hpp:rho}
\end{center}
\vskip -0.25in
\end{figure}

On the other hand, observe that the form of the natural gradient of $\omega_t$ have an intuitive semantic interpretation, which also extends to
the coordinate ascent variational message passing framework \citep{WinnBishop05} as shown by \citet{Mas16}. Specifically, using
the constant $\gamma$ as a threshold, we see that if the uninformed prior $p_u(\bmbeta_t)$ provides a better
fit to the variational posterior at time $t$ than the variational parameters
$\bmlambda_t$ from the previous time step ($\KL(q(\bmbeta_t\given \bmlambda_t), p_u(\bmbeta_t)) + \gamma < \KL(q(\bmbeta_t\given
\bmlambda_t),p_\delta(\bmbeta_t\given \bmlambda_{t-1}))$), then we will get a negative value for $\omega_t$ when performing 
coordinate ascent using Equation~(\ref{eq:natural_gradient}). This in turn implies that $\Exp_q[\rho]<0.5$ 
because $\Exp_q[\rho] = 1/(1- e^{-\omega_t}) - 1/ \omega_t$ (plotted in Figure \ref{fig:hpp:rho}), which means
that we have a higher degree of forgetting for past data. If $\omega_t > 0$ then $\Exp_q[\rho]>0.5$ and less past data is forgotten. Figure \ref{fig:hpp:rho} graphically illustrates this trade-off.

\subsection{The Multiple Hierarchical Power Prior Model}

The HPP model can immediately be extended to include multiple power priors $\rho_t^{(i)}$, one for each global parameter $\beta_i$. In this model the $\rho_t^{(i)}$'s are pair-wise independent. The latter ensures that optimizing the $\llb$ can be performed as above,
since the variational distribution for each $\rho_t^{(i)}$ can be updated independently of the other variational
distributions over $\rho_t^{(j)}$, for $j\neq i$. This extended model allows local model substructures to have
different forgetting mechanisms, thereby extending the expressivity of the model. We shall refer to this
extended model as a \emph{multiple hierarchical power prior} (MHPP) model.

\vekk{ 
\subsection{Variational Updating Continuation}

First of all, by instantiating the lower bound $\lb(q(\bmbeta_t, \bmz_t, \rho_t))$ in
Equation~(\ref{eq:likelihood_decomposition}) for the HPP model we obtain
\begin{equation}
  \label{eq:lower_bound}
\begin{split}
\lb&(q(\bmbeta_t, \bmz_t, \rho_t)) = \Exp_q [\ln p(\bmx_t,\bmZ_t\given \bmbeta_t)] \\
&+ \Exp_q[\ln
\hat{p}(\bmbeta_t\given \bmlambda_{t-1},\rho_t)]  \\
&+ \Exp_q[p(\rho_t\given \gamma)] - \Exp_q[\ln q(\bmZ_t\given
\bmphi_t)] \\
&- \Exp_q[q(\bmbeta_t\given \bmlambda_t)] - \Exp_q[q(\rho_t\given \omega_t)],  
\end{split}  
\end{equation}

where $\Exp_q[\ln \hat{p}(\bmbeta_t\given \bmlambda_{t-1},\rho_t)]$ is given by (ignoring the base measure), \textcolor{red}{as previously shown in Equation \ref{eq:PowerPrior}}:
\[
\Exp_q[(\rho_t\bmlambda_{t-1} + (1-\rho_t)\bmalpha_u)\bmt(\bmbeta_t) -  a_g(\rho_t\bmlambda_{t-1} + (1-\rho_t)\bmalpha_u)].
\]

From the expressions above we see that both $\nabla_{\bmlambda_t}\lb$ and $\nabla_{\bmphi}\lb$ only depend on $\rho_t$
through $E_q[\rho_t]$. Thus, updating the variational parameters $\bmlambda_t$ and $\bmphi_t$ in HPP models
can be done as for regular conjugate exponential models of the form in Figure~\ref{fig:plateModel}.

\textcolor{red}{I'm not sure about the last sentence making reference to model in Figure~\ref{fig:plateModel}.}

Finding $\nabla_{\omega_t}\lb$ is not as straightforward due to $a_g(\rho_t\bmlambda_{t-1} -
(1-\rho_t)\bmalpha_u)$ and \textcolor{red}{that we do not have a conjugate exponential model}. Instead we establish a lower bound $\llb \leq \lb$ and base the updating rule of
$\omega_t$ on $\llb$ rather than $\lb$. Specifically, we first exploit that $a_g(\cdot)$ is convex,
\[
a_g(\rho_t\bmlambda_{t-1} + (1-\rho_t)\bmalpha_u) \leq \rho_t a_g(\bmlambda_{t-1}) + (1-\rho_t)a_g(\bmalpha_u),  
\]
which combined with (\ref{eq:lower_bound}) gives
\begin{equation}
  \label{eq:lower_lower_bound}
\begin{split}
\llb &= \Exp_q [\ln p(\bmx_t,\bmZ_t\given \bmbeta_t)] \\
&+ \Exp_q[(\rho_t\bmlambda_{t-1} + (1-\rho_t)\bmalpha_u)\bmt(\bmbeta_t) -  \rho_t a_g(\bmlambda_{t-1}) \\
& - (1-\rho_t)a_g(\bmalpha_u)]
+ \Exp_q[p(\rho_t\given \gamma)] - \Exp_q[\ln q(\bmZ_t\given
\bmphi_t)] \\
&- \Exp_q[q(\bmbeta_t\given \bmlambda_t)] - \Exp_q[q(\rho_t\given \omega_t)] \leq \lb
\end{split}
\end{equation}

\begin{lem}
\label{lem:llbw}
  Let $\llb(\omega)$ be the part of $\llb$ that is non-constant wrt.\ $\omega$. If $p(\rho_t\given \gamma)$ and $q(\bmbeta_t\given
\bmlambda_t)$ belong to the same distribution family having the identity function as sufficient statistics
function, then 
\[
\begin{split}
\llb(\omega) &= \nabla_{\omega_t} a_g(\omega)(\Exp_q[\ln(p_\delta(\bmbeta_t\given \lambda_{t-1})) - \Exp_q[\ln
p_u(\bmbeta_t)]) \\
&+ \gamma \nabla_{\omega_t} a_g(\omega) - (\omega \nabla_{\omega_t} a_g(\omega) - a_g(\omega)) + \emph{cte} .
\end{split}
\]
\end{lem}
\begin{proof}
Firstly, by ignoring the terms in $\lb$ that do not involve $\omega$ and by exploiting the exponential family form of
$p_\delta(\bmbeta_t\given \lambda_{t-1})$ and $p_u(\bmbeta_t)$ we get
\[
\begin{split}
\llb(\omega) &= \Exp_q[\rho_t](\Exp_q[\ln(p_\delta(\bmbeta_t\given \lambda_{t-1})) - \Exp_q[\ln
p_u(\bmbeta_t)]) \\
&+ \Exp_q[p(\rho_t\given \gamma)]  - \Exp_q[q(\rho_t\given \omega)] + \emph{cte} .
\end{split}
\]
Since the sufficient statistics functions for $p(\rho_t\given \gamma)$ and $q(\bmbeta_t\given
\bmlambda_t)$ are assumed to be the identity function we have
\[
\begin{split}
\llb(\omega) &= \Exp_q[\rho_t](\Exp_q[\ln(p_\delta(\bmbeta_t\given \lambda_{t-1})) - \Exp_q[\ln
p_u(\bmbeta_t)]) \\
&+ \gamma \Exp_q[\rho_t] - (\omega \Exp_q[\rho_t] - a_g(\omega)) + \emph{cte} .
\end{split}
\]
Lastly, by using $\Exp_q[\bmt(\rho_t)] = \nabla_\omega a_g(\omega)$ we obtain the desired result.
\end{proof}

Based on Lemma~\ref{lem:llbw}, we can obtain a variational updating rule for $\omega$ by first finding the gradient
of $\llb$ wrt.\ $\omega$: 
\[
\nabla _\omega \llb = \nabla^2_\omega a_g(\omega)(\Exp_q[\ln(p_\delta(\bmbeta_t\given \lambda_{t-1})) - \ln
p_u(\bmbeta_t)] + \gamma - \omega ).
\]
From $\nabla _\omega \llb$ we find the \emph{natural gradient} \citep{sato2001online} (denoted $\hat\nabla
_\omega \llb$) by premultiplying
$\nabla _\omega \llb$ by the inverse of the Fisher information matrix, which for the exponential family corresponds to the inverse of the
Hessian of the log-normalizer: 
\[
\begin{split}
\hat\nabla _\omega \llb & = (\nabla^2_\omega a_g(\omega))^{-1}\nabla _\omega \llb \\
&=\Exp_q[\ln(p_\delta(\bmbeta_t\given \lambda_{t-1})) - \ln
p_u(\bmbeta_t)] + \gamma - \omega .
\end{split}
\]
  
We now have that $\hat\nabla _\omega \llb=0$ if and only if 
\[
\begin{split}
\omega = & \KL(q(\bmbeta_t\given \bmlambda_t), p_u(\bmbeta_t) \\
&- \KL(q(\bmbeta_t\given
\bmlambda_t),p_\delta(\bmbeta_t\given \bmlambda_{t-1})) + \gamma.
\end{split}
\]

Thus, the variational updating rule for $\omega$ attempts to \ldots

{\centerline {\bf Expand, possibly include figure}}

Collectively, the updating rules for the variational parameters in the HPP model are consistent in the sense that the regular updating rules for $\bmlambda_t$
and $\bmphi_t$ also maximizes the derived lower bound $\llb$ on which $\omega$ is based. This is formalized in
the following theorem.
\begin{thm}
  For a given HPP model, the variational gradient ascent updating rules for $\bmlambda_t$ and $\bmphi_t$ maximizes
  both $\lb$ and $\llb$:
  \[
      \hat\nabla_{\bmlambda_t}\lb  =   \hat\nabla_{\bmlambda_t}\llb \qquad
      \hat\nabla_{\bmphi_t}\lb =   \hat\nabla_{\bmphi_t}\llb \, .
  \]
\end{thm}
\begin{proof}
  The proof follows immediately by observing that 
  \[
  \begin{split}
  \llb - \lb = \Exp_q[&\rho_t a_g(\bmlambda_{t-1}) + (1-\rho_t)a_g(\bmalpha_u) \\
&+ a_g(\rho_t\bmlambda_{t-1} + (1-\rho_t)\bmalpha_u)   ]    
  \end{split}
\]
does not depend on neither $\bmlambda_t$ nor $\bmphi_t$.
\end{proof}

The HPP model can immediately be extended to include multiple power priors $\rho_t^{(i)}$, assuming that the
$\rho_t^{(i)}$'s are pair-wise independent. The latter ensures that optimizing the $\llb$ can be performed as above,
since the variational distribution for each $\rho_t^{(i)}$ can be updated independently of the other variational
distributions over $\rho_t^{(j)}$, for $j\neq i$. This extended model allows local model substructures to have
different forgetting mechanisms, thereby extending the expressivity of the model. We shall refer to this
extended model as a \emph{multiple hierarchical power prior} (MHPP) model.  

}



\section{Experiments}\label{sec:Experiments}

\subsection{Experimental Set-up}
In this section we will evaluate the  following methods:
\begin{itemize}[noitemsep,topsep=0pt,parsep=0pt,partopsep=0pt]
\item Streaming variational Bayes (SVB).
\item Four versions  of Population Variational Bayes (PVB)\footnote{We do not compare with SVI, because SVI is a special case of PVB when ${M}$ is equal to the total size of the stream.}: Population-size ${M}$ equal to the average size of each data-batch, or ${M}$ equal to a fixed value (${M}=1\,000$ in \secref{exp:artificial} and ${M}=10\,000$ in \secref{realData}). Learning-rate $\nu=0.1$ or $\nu=0.01$.
\item Two versions of SVB-PP: $\rho=0.9$ or $\rho=0.99$.
\item Two versions of SVB-HPP: A single shared $\rho$ (denoted SVB-HPP) or separate $\rho^{(i)}$ parameters (SVB-MHPP). 
\end{itemize}
The underlying variational engine is the VMP algorithm \cite{WinnBishop05} for all models; 
VMP was terminated after 100 iterations or if the relative increase in the lower bound fell below  $0.01\%$. 
All priors were uninformative, using either flat Gaussians, flat Gamma priors or uniform Dirichlet priors. We set $\gamma=0.1$ for the HPP priors. 
Variational parameters were randomly initialized using  the same seed for all  methods. 

\begin{figure}[t!]
\vskip -0.2in
\begin{center}
\begin{tabular}{ccc}
\includegraphics[width=9cm,height=5cm,keepaspectratio]{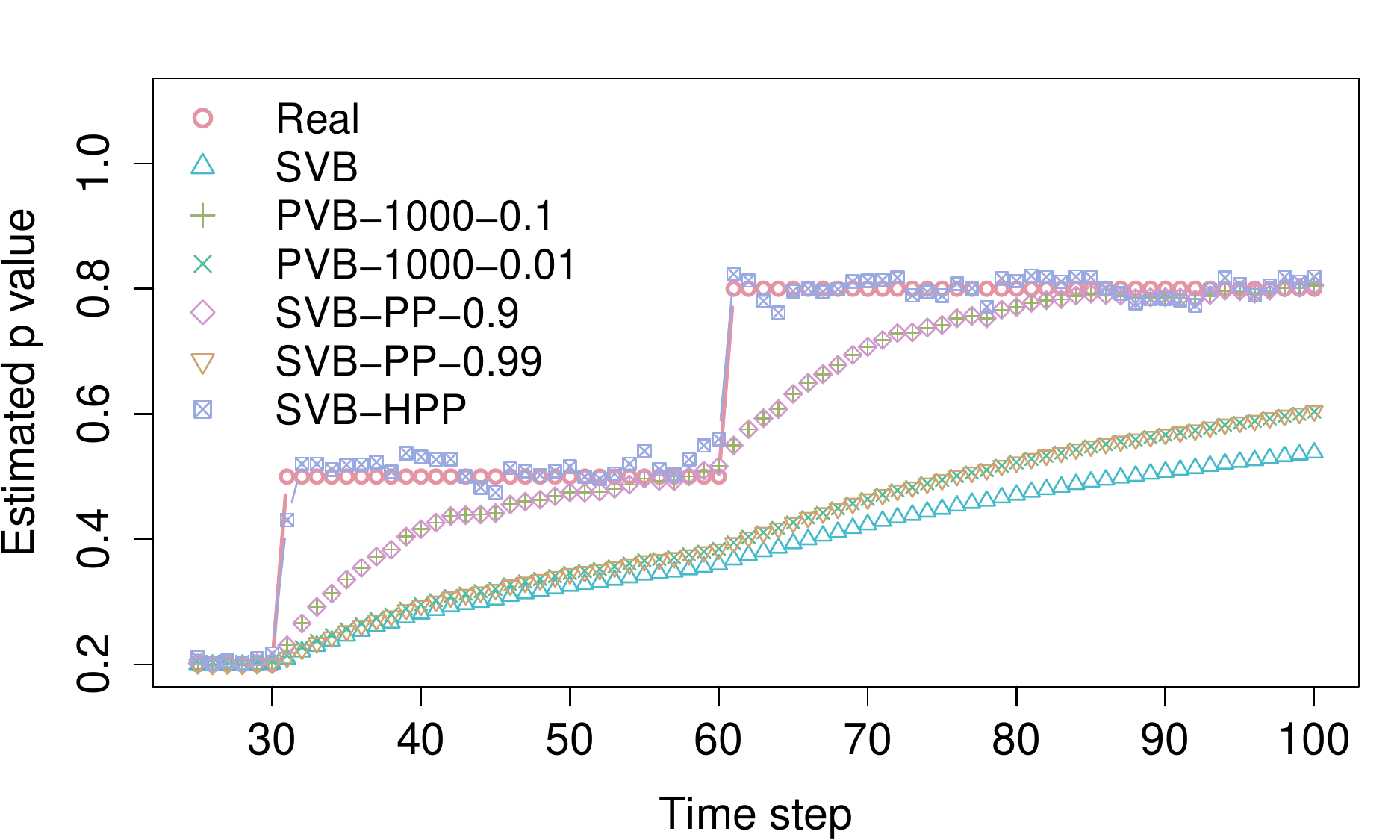}
\end{tabular}
\vskip -0.2in
\caption{ $E[\beta_t]$ in the Beta-Binomial model artificial data set} 
\label{fig:exp:artificialESS}
\end{center}
\vskip -0.25in
\end{figure}

\begin{figure*}[t!]
\vskip -0.1in
\begin{center}
\begin{tabular}{cc}
\includegraphics[width=8cm,height=4.5cm,keepaspectratio]{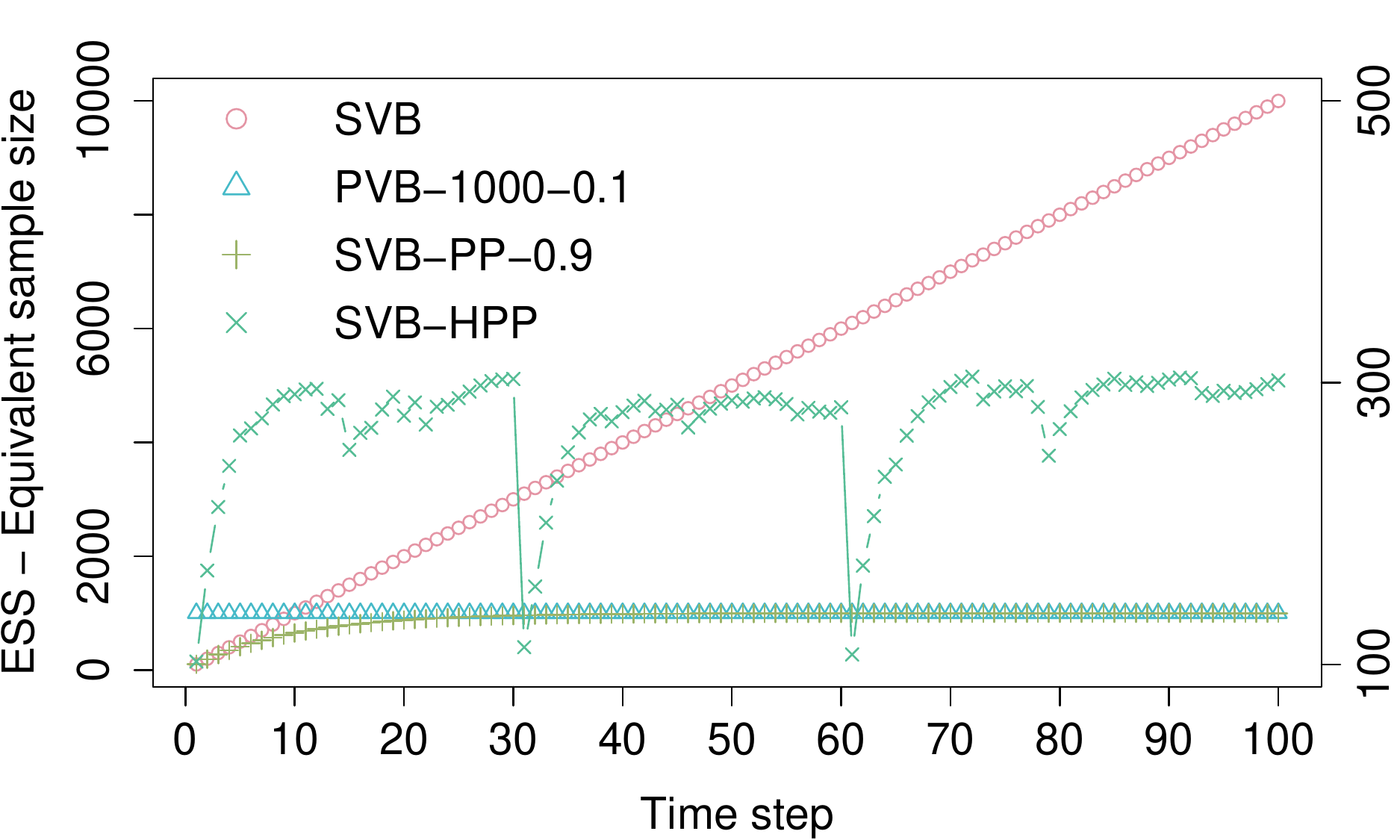}
&
\includegraphics[width=8cm,height=4.5cm,keepaspectratio]{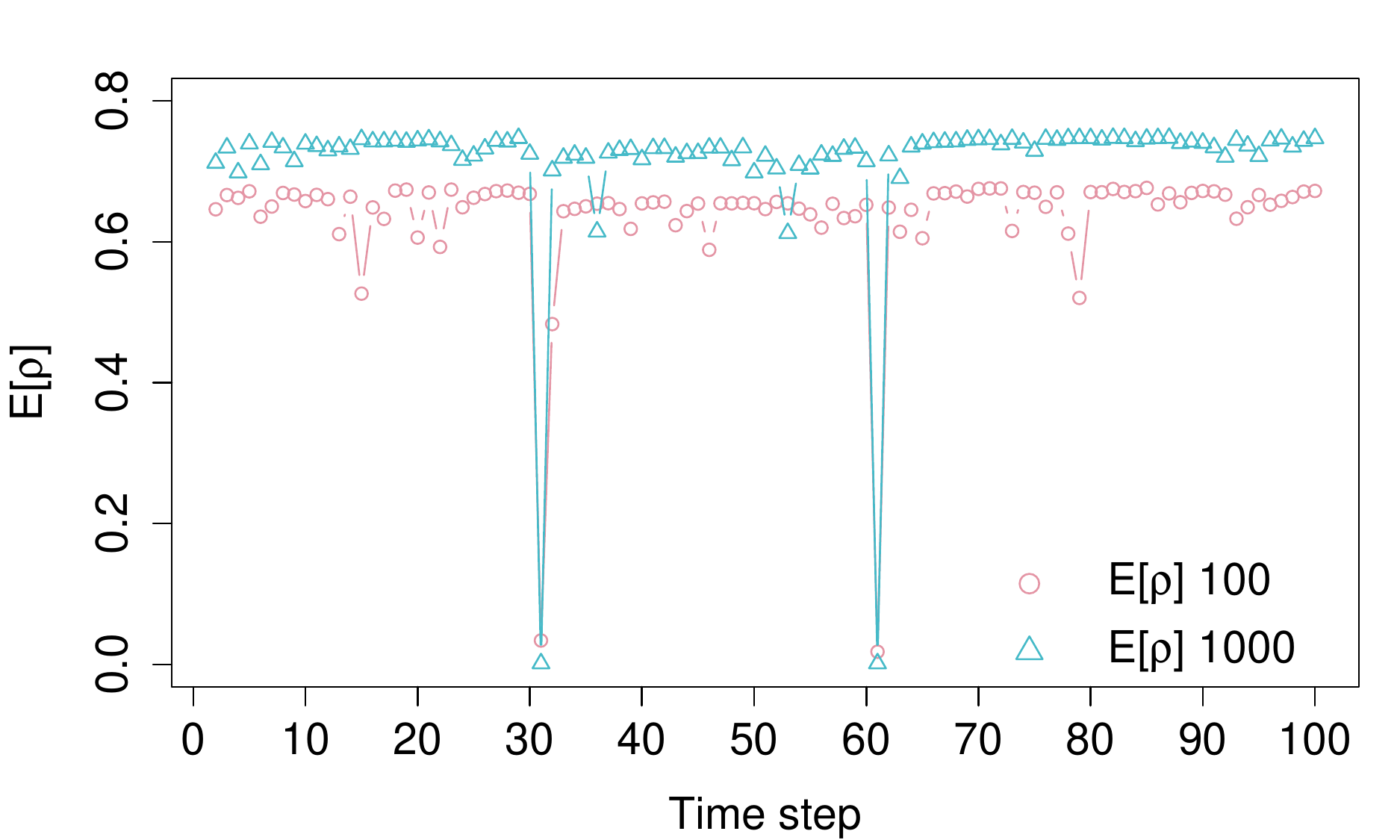}\\
\end{tabular}
\vskip -0.1in
\caption{$ESS_t$ (left) and $\mathbb{E}_q[\rho_t]$ (right) in the Beta-Binomial model artificial data set}
\label{fig:exp:artificialParemeter}
\end{center}
\vskip -0.2in
\end{figure*}

\subsection{Evaluation using an Artificial Data Set}
\label{sec:exp:artificial}

First, we illustrate the behavior of the different approaches in a controlled experimental setting: 
We produced an artificial data stream by generating 100 samples (i.e., $|\bmx_t|=100$) from a Binomial distribution at each time step. 
We artificially introduce concept drift by changing the parameter $p$ of the Binomial distribution: $p=0.2$ for the first 30 time steps, then $p=0.5$ for the following 30 time steps, and finally $p=0.8$  for the last 40 time steps. The data stream was modelled using a Beta-Binomial model.  
 
\textbf{Parameter Estimation:} \figref{exp:artificialESS} shows the evolution of $\mathbb{E}_q[\beta_t]$ for the different methods. 
We recognize that SVB simply generates a running average of the data, as it is not able to adapt to the concept drift. 
The results from PVB depend heavily on the learning rate $\nu$, where the higher learning rate, which results in the more aggressive forgetting, works better in this example. 
Recall, though, that $\nu$  needs to be hand-tuned to achieve an optimal performance. 
As expected,  the choice of ${M}$  does not have an impact, because the present model has no local hidden variables (cf.\ \secref{RelatedWork}). 
SVB-PP produces results almost identical to PVB when $\rho$ matches the learning rate of PVB (i.e., $\rho = 1-\nu$). 
Finally, SVB-HPP provides the best results, almost mirroring the true model.

\textbf{Equivalent Sample Size (ESS):} \figref{exp:artificialParemeter} (left) gives the evolution of the equivalent sample size, $ESS_t$, for the different methods \footnote{For this model, ESS is simply computed by summing up the components of the $\lambda_t$ defining the Beta posterior.}. 
The ESS of PVB is always given by the constant ${M}$. 
For SVB, the ESS monotonically increases as more data is seen, while SVB-PP exhibits convergence to the limiting value computed in Equation (\ref{eq:ESSLimit}). 
A different behaviour is observed for SVB-HPP:  It is automatically adjusted. 
Notice that the values for this model is to be read off the alternative $y$-axis. We can detect the the concept drift, by identifying where the ESS rapidly declines. 

\textbf{Evolution of Expected Forgetting factor:} In \figref{exp:artificialParemeter} (right) the series denoted ``$E[\rho]-100$'' shows the evolution of $\mathbb{E}_q [\rho_t]$ for the artificial data set.  Notice how the model clearly identifies abrupt concept drift at time steps $t=30$ and $t=60$. 
The series denoted ``$E[\rho]-1000$'' illustrates the evolution of the parameter when we increase the batch size to 1000 samples. 
We recognize a more confident assessment about the absence of concept drift as more data is made available.

\subsection{Evaluation using Real Data Sets}\label{sec:realData}



\subsubsection{Data and Models}

For this evaluation we consider three real data sets from different domains:


\textbf{Electricity Market}  \cite{harries1999splice}: The data set describes the electricity market of two Australian states. It contains 45312 instances of 6 attributes, including a class label comparing the change of the electricity price related to a moving average of the last 24 hours.  
Each instance in the data set represents 30 minutes of trading; during our analysis we created batches such that $\bmx_t$ contains all information associated with month $t$. 

The data is analyzed using a Bayesian linear regression model. The binary class label is assumed to follow a Gaussian distribution in order to fit within the conjugate model class. Similarly, the marginal densities of the predictive attributes are also assumed to be Gaussian. 
The regression coefficients are given Gaussian prior distributions, and the variance is given a Gamma prior. 
Note that the overall distribution does not fall inside the conditional conjugate exponential family \cite{HoffmanBleiWangPaisley13}, hence PVB cannot be applied here, because lower-bound's gradient cannot be computed in closed-form.

\textbf{GPS}   \cite{GPS2,GPS1,GPS3}: This data set contains $17\,621$ GPS trajectories (time-stamped $x$ and $y$ coordinates), totalling more than 4.5 million observations. To reduce the data-size we kept only one out of every ten measurements. We grouped the data so that $\bmx_t$ contains all data collected during hour $t$ of the day, giving a total of $24$ batches of this stream.

Here we employ a model with one independent Gaussian mixture model per day of the week,  each mixture with 5 components. This enables us to track changes in the users' profiles across hours of the day, and also to monitor how the changes are affected by the day of the week.



\textbf{Finance} (reference withheld): The data contains monthly aggregated information about the financial profile of around $50\,000$  customers over 62 (non-consecutive) months. Three attributes were extracted per customer, in addition to a class-label telling whether or not the customer will default within the next 24 months.

We fit a na\"{\i}ve Bayes model to this data set, where the distribution at the leaf-nodes is 5-component mixture of Gaussians distribution. 
The distribution over the mixture node is shared by all the attributes, but not between the two classes of customers.

A detailed description of all the models, including their structure and their variational families, is given at the supplementary material.

\begin{table*}[htb!]
\vskip 0.15in
\begin{center}
\begin{scriptsize}
\begin{sc}
\begin{tabular}{lccccccccc}
\hline
Data set & SVB & PVB & PVB & PVB & PVB & SVB-PP & SVB-PP & SVB-HPP & SVB-MHPP \\
			 &        &  $M=10k$ & $M=10k$ & $M=|\bmx_t|$  & $M=|\bmx_t|$ & $\rho = 0.9$  & $\rho = 0.99$ & & \\
 			 &        &  $\nu = 0.1$ & $\nu = 0.01$  & $\nu = 0.1$ & $\nu = 0.01$ &   &  & &\\
\hline
Electricity & -44.91 &  &  &  &  & -43.92 & -44.80 & -40.06 & \textbf{-40.03}\\
GPS & -1.93 & -2.03 & -2.72 & -1.88 & -2.42 & -1.89 & -1.92 & -1.86 & \textbf{-1.74}\\
Finance & -19.84 & -22.29 & -22.57 & -21.81 & -22.07 & \textbf{-19.05} & -19.78 & -19.83 & -19.40\\
\hline
\end{tabular}
\end{sc}
\end{scriptsize}
\end{center}
\vskip -0.1in
\caption{Aggregated Test Marginal Log-Likelihood.\protect{\label{table:exp:totalMLL}}}
\end{table*}


\begin{figure*}[htb!]
\begin{center}
\begin{tabular}{cc}
\includegraphics[width=8cm,height=4.5cm,keepaspectratio]{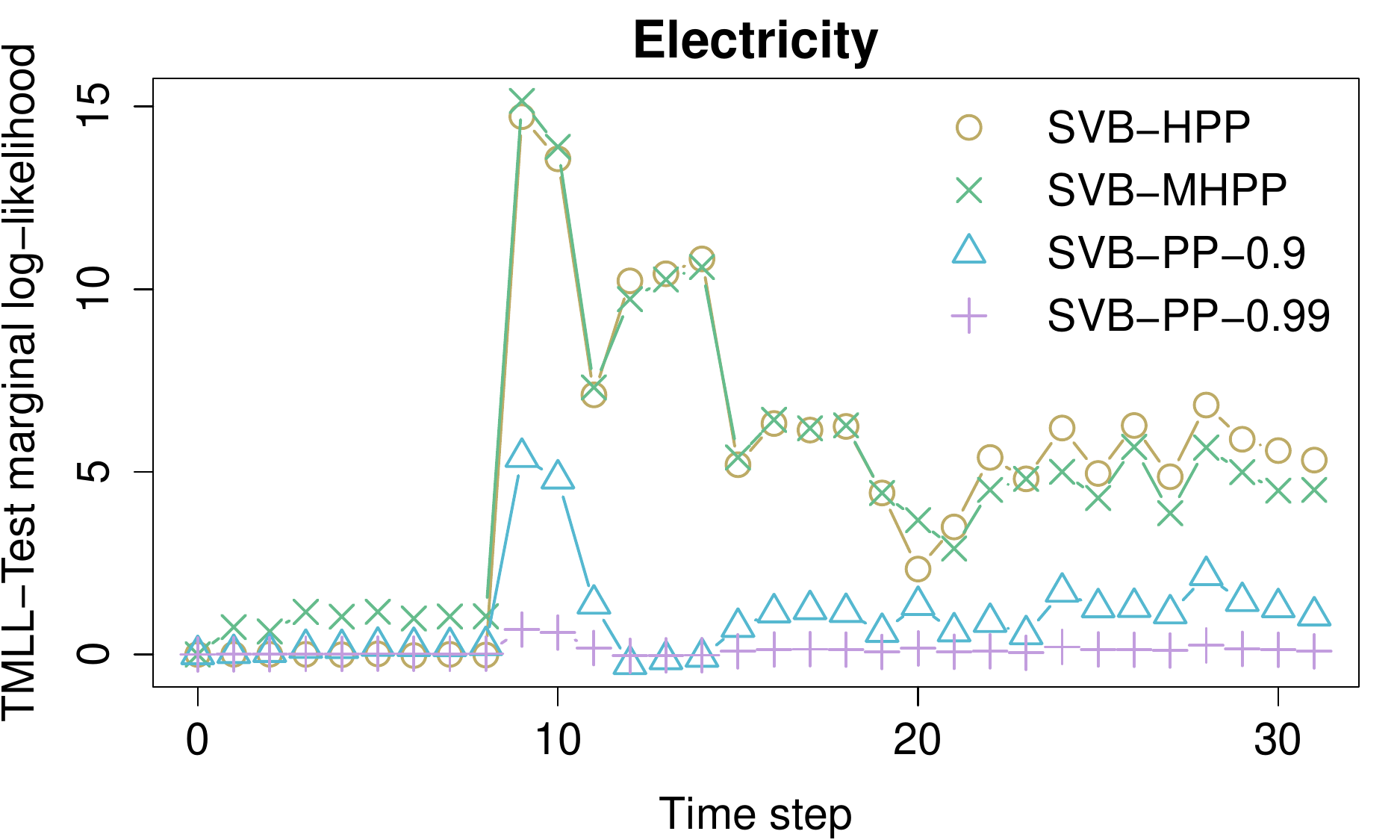}&
\includegraphics[width=8cm,height=4.5cm,keepaspectratio]{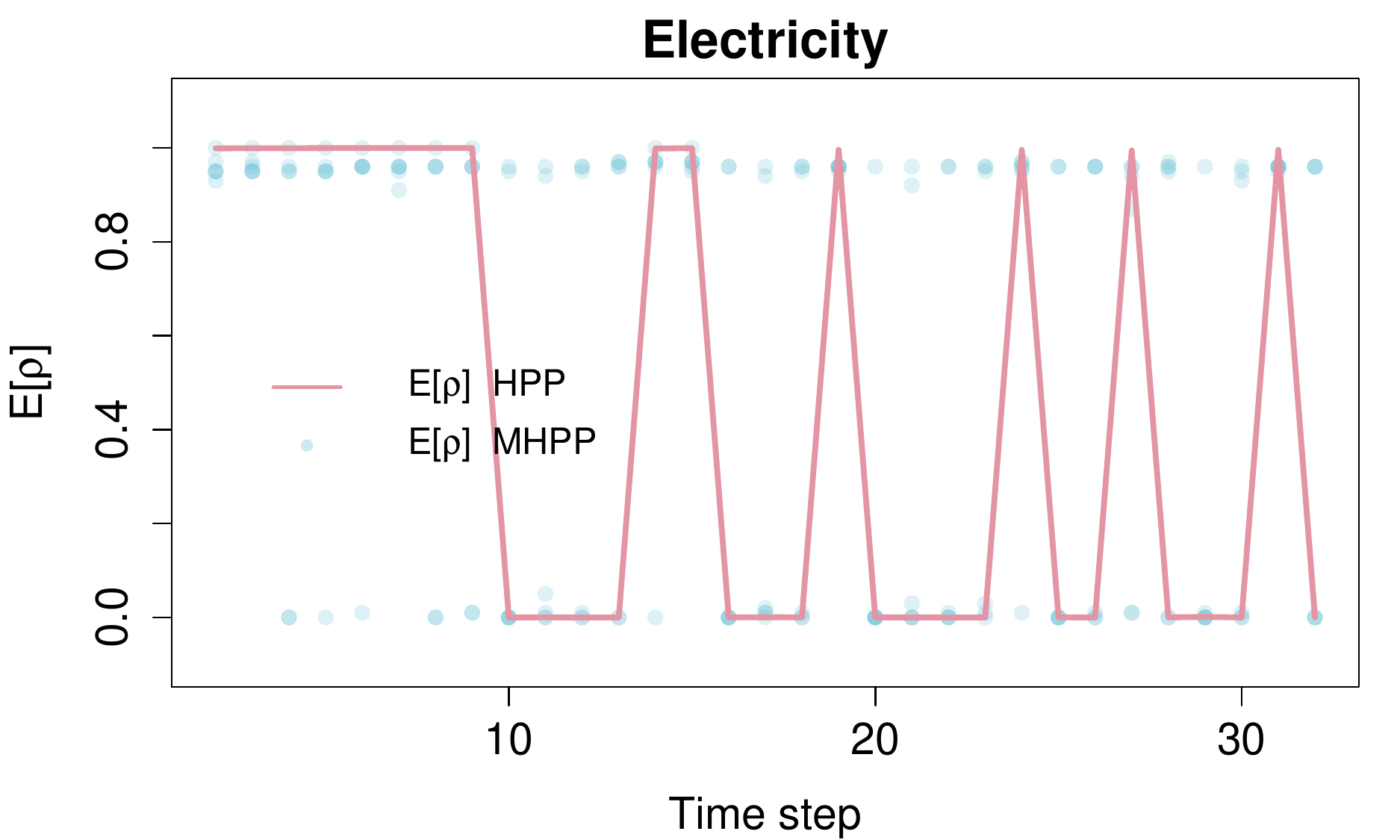}\\

\includegraphics[width=8cm,height=4.5cm,keepaspectratio]{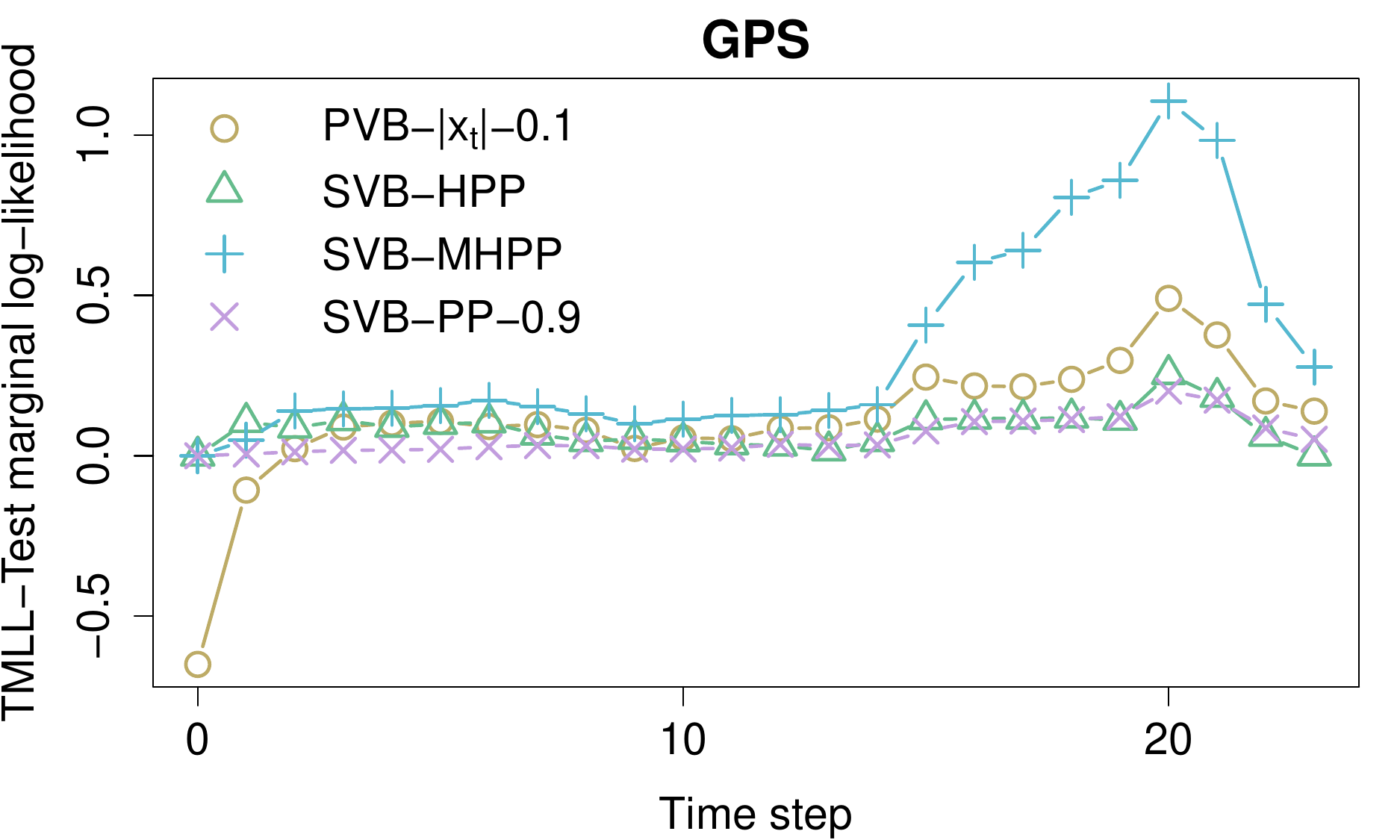}&
\includegraphics[width=8cm,height=4.5cm,keepaspectratio]{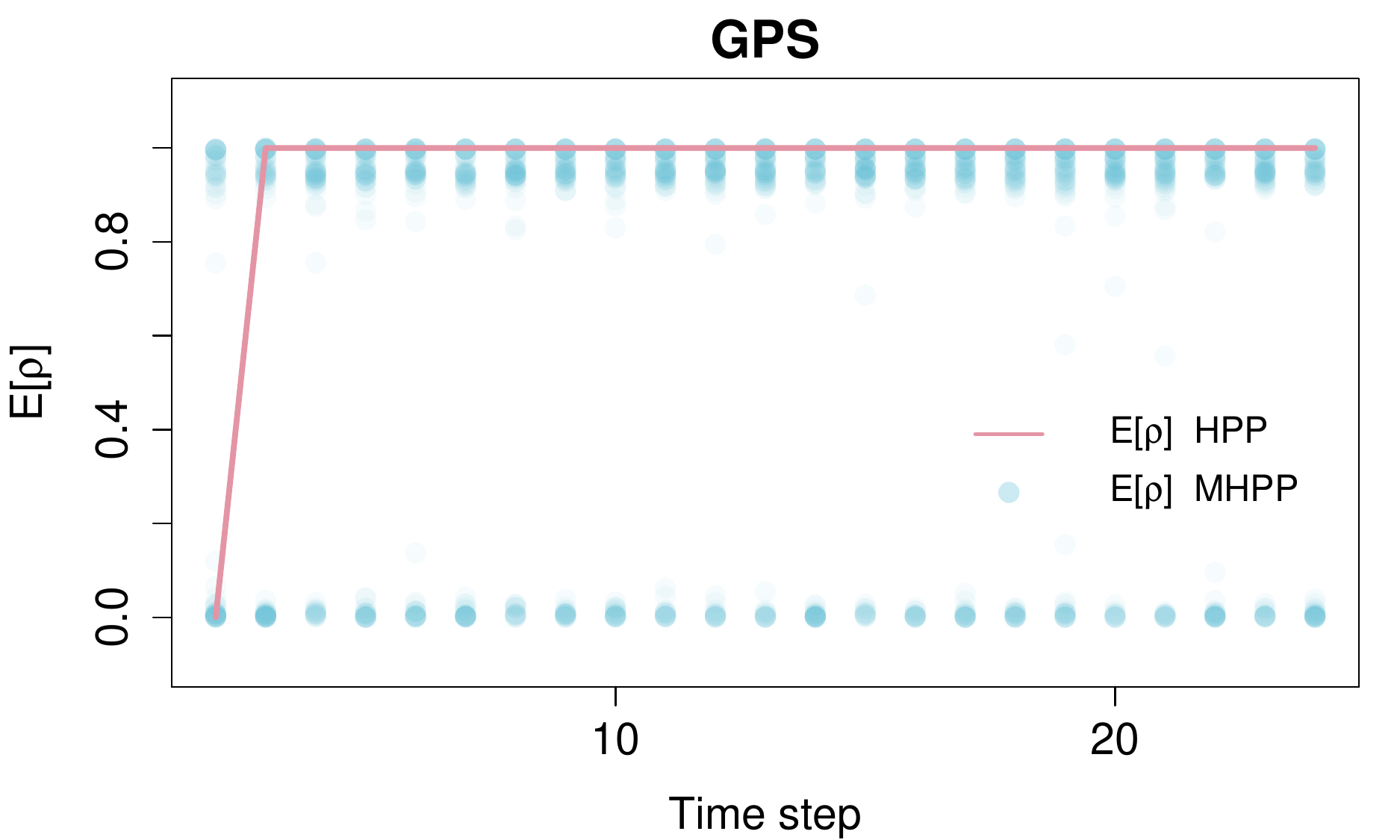}\\

\includegraphics[width=8cm,height=4.5cm,keepaspectratio]{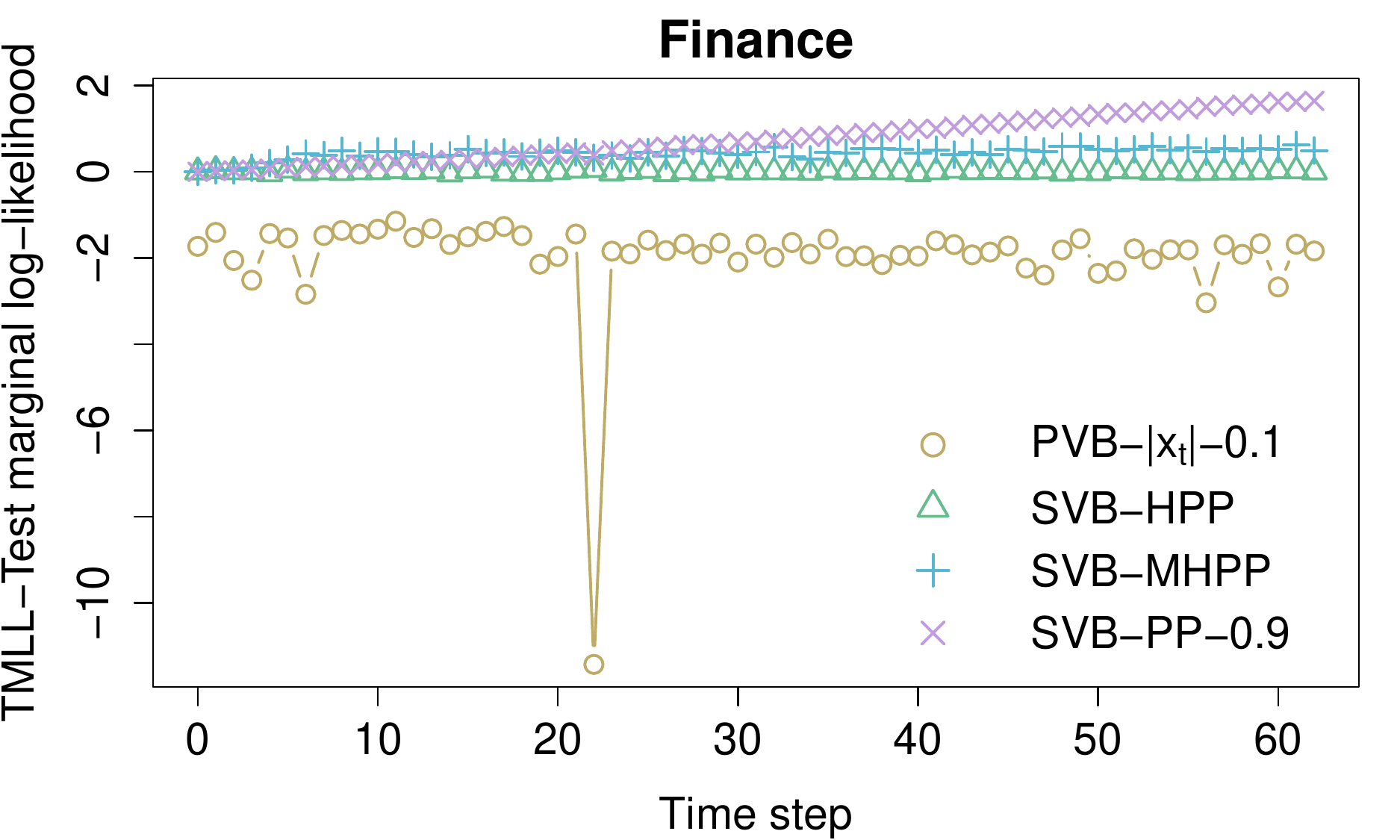}&
\includegraphics[width=8cm,height=4.5cm,keepaspectratio]{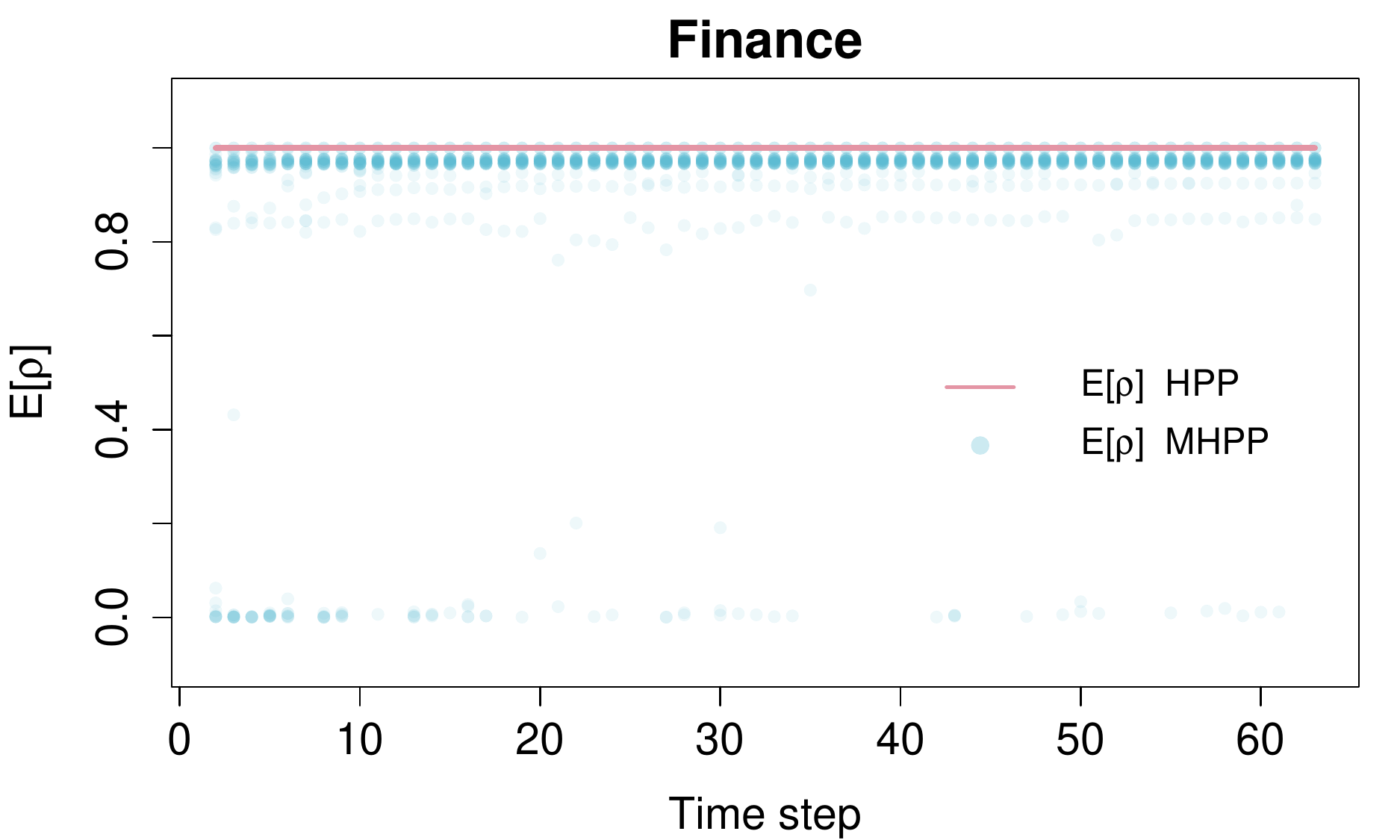}\\
(a) TMLL$_t$ improvement over SVB  & (b) $\mathbb{E}_q[\rho_t]$ for SVB-HPP and SVB-MHPP\\
\end{tabular}
\caption{Results of the competing methods for the three real-life data sets. See text for discussion.}
\label{fig:exp:elec}
\end{center}
\vskip -0.2in
\end{figure*}

%

\subsubsection{Evaluation and Discussion}

To evaluate the different methods discussed, we look at the test marginal log-likelihood (TMLL). 
Specifically, each data batch   is randomly split in a train data set, $\bmx_t$, and a test data set, \testData, containing two thirds and one third of the data batch, respectively. Then, TMLL$_t$ is computed as  
$\mbox{TMLL}_t = \frac{1}{|\testData|}\int p(\testData,{\bmz_t}|\bmbeta_t)p(\bmbeta_t|\bmx_t) d\bmz_t d\bmbeta_t.$ 
\figref{exp:elec} (left) shows for each method the difference between its TMLL$_t$ and that obtained by SVB (which is considered the baseline method). 
To improve readability, we only plot the results of the best performing method inside each group of methods. 
The right-hand side of \figref{exp:elec} shows the development of  $\Exp_q[\rho_t]$ over time for  SVB-HPP and SVB-MHPP. 
For SVB-HPP we only have one $\rho_t$-parameter, and its value is given by the solid line. 
SVB-MHPP utilizes one $\rho^{(i)}$  for each variational parameter.\footnote{The numbers of variational parameters are 14, 78 and 33 for the Electricity, GPS and Financial model, respectively.}
In this case, we plot $\Exp_q[\rho^{(i)}_{t}]$ at each point in time to indicate the variability between the different estimates throughout the series.
Finally, we  compute each method's aggregated test marginal log-likelihood measure $\sum_{t=1}^T \mbox{TMLL}_t$, and report these values in Table \ref{table:exp:totalMLL}. 

For the electricity data set, we can see that  the two proposed methods (SVB-HPP and SVB-MHPP) perform best. 
All models are comparable during the first nine months, which is a period where our models detect no or very limited concept drift (cf.\ top right plot or \figref{exp:elec}).
However, after this period, both SVB-HPP and SVB-MHPP detects substantial drift, and is able to adapt better than the other methods, which appear unable to adjust to the complex concept drift structure in the latter part of the data. 
SVB-HPP and SVB-MHPP continue to behave at a similar level, mainly because when drift happens it  typically includes a high proportion of the parameters of the model.  

For the GPS data set, we can observe how the SVB-MHPP is superior to the rest of the methods, particularly towards  the end of the series. 
When looking at \figref{exp:elec} (middle right panel), we can see  that  a significative proportion of the model parameters are drifting (i.e., $\Exp_q[\rho^{(i)}_{t}]\leq 0.05$) at all times, while another proportion of the parameters show a quite stable behavior ($\rho$-values  above $0.9$). 
This complex pattern is not captured well by SVB-HPP, which ends up assuming no concept drift after the initial time-step. 

The financial data set shows a different behavior. During the first months, SVB-MHPP slightly outperforms the rest of the approaches, but after month 30, 
SVB-PP with $\rho=0.9$ is superior, with SVB-MHPP second. 
Looking at the $E[\rho^{(i)}_t]$-values of SVB-MHPP, we observe that there is significant concept drift in some of the parameters over the first few months. 
However, only a few parameters exhibit noteworthy drift after the first third of the sequence. 
Apparently, the simple SVB-PP approach has the upper hand when the drift is constant and fairly limited, at least when the optimal forgetting factor $\rho$ has been identified.

We conclude this section by highlighting that the performance of  SVB-PP and PVB depend  heavily on the hyper-parameters of the model, cf.\ Table \ref{table:exp:totalMLL}. 
As an example, consider SVB-PP for the financial data set. While it was the best overall with $\rho=0.9$, it is inferior to SVB-MHPP if $\rho=0.99$.
Similarly, PVB's performance is sensitive both to $\nu$ (see in particular the results for the GPS data) and ${M}$ (financial data). 
These hyper-parameters are hard to fix, as their optimal values depend on data characteristics (see \citet{BroderickBoydWibisonoWilsonJordan13,McInerneyRanganathBlei15} for similar conclusions). 
We therefore believe that the fully Bayesian formulation is an important strong point of our approach.

\section{Conclusions and Future Work}\label{sec:ConclusionsAndFutureWork}

We have introduced a new class of Bayesian models for streaming data, able to capture changes in
the underlying generative process. Unlike existing solutions to this problem, aimed at modeling slowly
changing processes, our proposal is able to handle both abrupt and gradual concept drift following
a Bayesian approach. The new model accounts for the dynamics of the data stream by assuming that only
the global parameters evolve over time. We introduce the so-called hierarchical power priors,
where a prior on the learning rate is given allowing it to adapt to the data stream.
We have addressed the complexity of the underlying inference tasks by
developing an approximate variational inference scheme that optimizes a novel lower bound of the
likelihood function.

As future work we aim to provide a sound approach to semantically characterize concept drift 
by inspecting the $\Exp[\rho^{(i)}_t]$ values provided by SVB-MHPP.



\section*{Acknowledgements} 
This work was partly carried out as part of the AMIDST project. AMIDST has received
funding from the European Union's Seventh Framework Programme for research,
technological development and demonstration under grant agreement no 619209.
Furthermore, this research has been partly funded by the Spanish Ministry of Economy and Competitiveness, 
through projects TIN2015-74368-JIN, TIN2013-46638-C3-1-P,  TIN2016-77902-C3-3-P and by ERDF funds.

\bibliography{bibliography}
\bibliographystyle{icml2017}

\end{document}


\twocolumn[
\icmltitle{Supplementary Material for Bayesian models of Data Streams with HPPs}




\begin{icmlauthorlist}
\icmlauthor{Andr\'{e}s Masegosa}{ual,ntnu}
\icmlauthor{Thomas D. Nielsen}{aau}
\icmlauthor{Helge Langseth}{ntnu}
\icmlauthor{Dar\'{\i}o Ramos-L\'{o}pez}{ual}
\icmlauthor{Antonio Salmer\'{o}n}{ual}
\icmlauthor{Anders L. Madsen}{aau,hug}
\end{icmlauthorlist}

\icmlaffiliation{ual}{Department of Mathematics, Unversity of Almer\'{\i}a, Almer\'{\i}a, Spain}
\icmlaffiliation{aau}{Department of Computer Science, Aalborg University, Aalborg, Denmark}
\icmlaffiliation{ntnu}{Department of Computer and Information Science, Norwegian University of Science and Technology, 
Trondheim, Norway}
\icmlaffiliation{hug}{Hugin Expert A/S, Aalborg, Denmark}
\icmlcorrespondingauthor{Andr\'{e}s Masegosa}{andresmasegosa@ual.es}

\icmlkeywords{data streams, Bayesian model, adaptivity, power priors, concept drift}

\vskip 0.3in
]



\printAffiliationsAndNotice{}  

\appendix

\section{Proof of Theorem 1 and Lemma 2}

\begin{proof}[Proof of Theorem 1]
  In the specification of $\lb$ we have that $\Exp_q[\ln \hat{p}(\bmbeta_t\given \bmlambda_{t-1},\rho_t)]$
  (defined in Equation~(7)) can be expanded as (ignoring the base measure) :
\[
\Exp_q[(\rho_t\bmlambda_{t-1} + (1-\rho_t)\bmalpha_u)\bmt(\bmbeta_t) -  a_g(\rho_t\bmlambda_{t-1} + (1-\rho_t)\bmalpha_u)].
\]
Since $a_g$ is convex we have
\[
a_g(\rho_t\bmlambda_{t-1} + (1-\rho_t)\bmalpha_u) \leq \rho_t a_g(\bmlambda_{t-1}) + (1-\rho_t)a_g(\bmalpha_u),  
\]
which combined with Equation~(10) gives
\[
\begin{split}
&\Exp_q [\ln p(\bmx_t,\bmZ_t\given \bmbeta_t)] + \Exp_q[(\rho_t\bmlambda_{t-1} +
(1-\rho_t)\bmalpha_u)\bmt(\bmbeta_t) \\
&-  \rho_t a_g(\bmlambda_{t-1})  - (1-\rho_t)a_g(\bmalpha_u)]
+ \Exp_q[p(\rho_t\given \gamma)] \\
&- \Exp_q[\ln q(\bmZ_t\given
\bmphi_t)] - \Exp_q[q(\bmbeta_t\given \bmlambda_t)] - \Exp_q[q(\rho_t\given \omega_t)] \leq \lb.
\end{split}
\]
Lastly, by exploiting the mean field factorization of $q$ and using the exponential family form of
$p_\delta(\bmbeta_t\given \lambda_{t-1})$ and $p_u(\bmbeta_t)$ we get the desired result.
\end{proof}

\begin{proof}[Proof of Lemma 2]
  Firstly, by ignoring the terms in $\llb$ (Equation~(11)) that do not involve $\omega_t$
  we get 
\[
\begin{split}
\llb(\omega_t) &= \Exp_q[\rho_t](\Exp_q[\ln(p_\delta(\bmbeta_t\given \lambda_{t-1})) - \Exp_q[\ln
p_u(\bmbeta_t)]) \\
&+ \Exp_q[p(\rho_t\given \gamma)]  - \Exp_q[q(\rho_t\given \omega_t)] .
\end{split}
\]
Assuming that the sufficient statistics function $\bmt(\rho_t)$ for $p(\rho_t\given \gamma)$ and $q(\bmbeta_t\given
\bmlambda_t)$ is the identity function ($\bmt(\rho_t)=\rho_t$) we have
\[
\begin{split}
\llb(\omega_t) &= \Exp_q[\rho_t](\Exp_q[\ln(p_\delta(\bmbeta_t\given \lambda_{t-1})) - \Exp_q[\ln
p_u(\bmbeta_t)]) \\
&+ \gamma \Exp_q[\rho_t] - (\omega_t \Exp_q[\rho_t] - a_g(\omega_t)) + \emph{cte} .
\end{split}
\]
Using $\Exp_q[\bmt(\rho_t)] = \Exp_q[\rho_t] = \nabla_{\omega_t} a_g(\omega_t)$ we get
\[
\begin{split}
\llb(\omega_t) &= \nabla_{\omega_t} a_g(\omega_t)(\Exp_q[\ln(p_\delta(\bmbeta_t\given \lambda_{t-1})) - \Exp_q[\ln
p_u(\bmbeta_t)]) \\
&+ \gamma \nabla_{\omega_t} a_g(\omega_t) - (\omega_t \nabla_{\omega_t} a_g(\omega_t) - a_g(\omega_t)). 
\end{split}
\]
and thereby
\[
\nabla _{\omega_t} \llb = \nabla^2_{\omega_t} a_g(\omega_t)(\Exp_q[\ln(p_\delta(\bmbeta_t\given \lambda_{t-1})) - \ln
p_u(\bmbeta_t)] + \gamma - \omega_t ).
\]
We can now find the natural gradient by premultiplying
$\nabla _{\omega_t} \llb$ by the inverse of the Fisher information matrix, which for the exponential family corresponds to the inverse of the
Hessian of the log-normalizer: 
\[
\begin{split}
\hat\nabla _{\omega_t} \llb & = (\nabla^2_{\omega_t} a_g(\omega_t))^{-1}\nabla _{\omega_t} \llb \\
&=\Exp_q[\ln(p_\delta(\bmbeta_t\given \lambda_{t-1})) - \ln
p_u(\bmbeta_t)] + \gamma - \omega_t .
\end{split}
\]
Lastly, by introducing $q(\bmbeta_t\given \bmlambda_t)-q(\bmbeta_t\given \bmlambda_t)$ inside the expectation
we get the difference in Kullbach-Leibler divergence $\KL(q(\bmbeta_t\given \bmlambda_t), p_u(\bmbeta_t)) - \KL(q(\bmbeta_t\given
\bmlambda_t),p_\delta(\bmbeta_t\given \bmlambda_{t-1}))$. 
\end{proof}

\section{Experimental Evaluation}

\subsection{Probabilistic Models}

We provide a (simplified) graphical description of the probabilistic models used in the experiments. We also detail the distributional assumptions of the parameters, which are then used to define the variational approximation family.

\subsubsection*{Electricity Model}


\begin{figure}[htb!]
	\begin{center}
		\scalebox{0.9}{
			\begin{tikzpicture}
			\nc{\nsize}{0.9cm}
			
			\node[latent, minimum size=\nsize]  (x1)   {${x}_{1,t}$}; %
			
			\node[latent, right=.5 of x1, minimum size=\nsize]  (x2)   {${x}_{2,t}$}; %
			
			\node[latent, right=.5 of x2, minimum size=\nsize]  (x3)   {${x}_{3,t}$}; %
			
			\node[latent, below=1 of x2, minimum size=\nsize]  (y)   {${y}_{t}$}; %
			\edge{x1}{y};
			\edge{x2}{y};
			\edge{x3}{y};
			
			\end{tikzpicture}
			
		}
	\end{center}
	\label{fig_DAG_electricity}
\end{figure}

\begin{align*}
    (\mu_i, \gamma_i) & \sim {NormalGamma}(1,1,0,1e-10)
\\ \gamma & \sim {Gamma}(1,1)
\\  b_i & \sim \mathcal{N}(0, +\infty) 
\\  x_{i,t} & \sim \mathcal{N}(\mu_i, \gamma_i)
\\  y_t & \sim \mathcal{N} \left( b_0 + \sum_{i} b_i x_{i,t}, \gamma \right) 
\end{align*}

\subsubsection*{GPS Model}


\begin{figure}[htb]
	\begin{center}
		\scalebox{0.9}{
			\begin{tikzpicture}
			\nc{\nsize}{0.9cm}
			
			\node[latent, minimum size=\nsize]  (z)   {${z}_{t}$}; %
			
			\node[latent, above=.5 of z, minimum size=\nsize]  (day)   {${Day}_{t}$}; %
			\edge{day}{z};
						
			\node[latent, below left=1 of z, minimum size=\nsize]  (x)   {${x}_{t}$}; %
			\edge{z}{x};
			\edge{day}{x};
			
			\node[latent, below right=1 of z, minimum size=\nsize]  (y)   {${y}_{t}$}; %
			\edge{z}{y};
			\edge{day}{y};			
			
			\end{tikzpicture}
		}
	\end{center}
	\label{fig_DAG_GPS}
\end{figure}

\begin{align*}
    p & \sim {Dirichlet}(1,\hdots,1)
\\ p_k & \sim {Dirichlet}(1,\hdots,1)
\\ (\mu^{(x)}_{j,k}, \gamma^{(x)}_{j,k})& \sim {NormalGamma}(1,1,0,1e-10)
\\ (\mu^{(y)}_{j,k}, \gamma^{(y)}_{j,k})& \sim {NormalGamma}(1,1,0,1e-10)
\\ Day_t &\sim {Multinomial}(p)
\\ \left( z_t | Day_t=k \right) &\sim {Multinomial}(p_k)
\\ \left( x_t | z_t=j, Day_t=k \right) &\sim \mathcal{N} ( \mu^{(x)}_{j,k}, \gamma^{(x)}_{j,k})
\\ \left( y_t | z_t=j, Day_t=k \right) &\sim \mathcal{N} ( \mu^{(y)}_{j,k}, \gamma^{(y)}_{j,k}) 
\end{align*}

\subsubsection*{Financial Model}


\begin{figure}[htb!]
	\begin{center}
		\scalebox{0.9}{
			\begin{tikzpicture}
			\nc{\nsize}{0.9cm}
			
			\node[latent, minimum size=\nsize]  (z)   {${z}_{t}$}; %
			
			\node[latent, above=.5 of z, minimum size=\nsize]  (default)   {${Default}_{t}$}; %
			\edge{default}{z};
			
			\node[latent, below left = 2 of z, minimum size=\nsize]  (x1)   {${x}_{1,t}$}; %
			\edge{z}{x1};
			\edge{default}{x1};
			
			\node[latent, below = 1.14 of z, minimum size=\nsize]  (x2)   {${x}_{2,t}$}; %
			\edge{z}{x2};
			\draw [->] (default) to [out=250,in=150] (x2);
			
			\node[latent, below right = 2 of z, minimum size=\nsize]  (x3)   {${x}_{3,t}$}; %
			\edge{z}{x3};
			\edge{default}{x3};				
			
			\end{tikzpicture}
		}
	\end{center}
	\caption{Simplified DAG for the financial model}
	\label{fig_DAG_financial}
\end{figure}
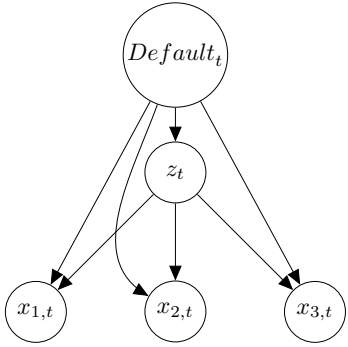

\begin{align*}
    p & \sim {Dirichlet}(1,\hdots,1)
\\ p_k & \sim {Dirichlet}(1,\hdots,1)
\\  (\mu_{i; j,k}, \gamma_{i; j,k}) & \sim {NormalGamma}(1,1,0,1e-10)
\\ Default_t & \sim Binomial(p)
\\ \left( z_t | Default_t=k \right)  & \sim {Multinomial}(p_k)
\\ \left( x_{i,t} | z_t=j, Day_t=k \right) &  \sim \mathcal{N} ( \mu_{i; j,k}, \gamma_{i; j,k})
\end{align*}

%
%
%

\subsection{Real Life Data Sets}
In the experimental section of the original paper, we plot the relative values for the $TMLL_t$ measure with respect to the SVB method. Here, we provides the plots of the absolute values of the $TMLL_t$ series for the different methods studied in the paper.

\begin{figure}[htb!]
	\begin{center}
		\includegraphics[width=8cm,height=4.5cm,keepaspectratio]{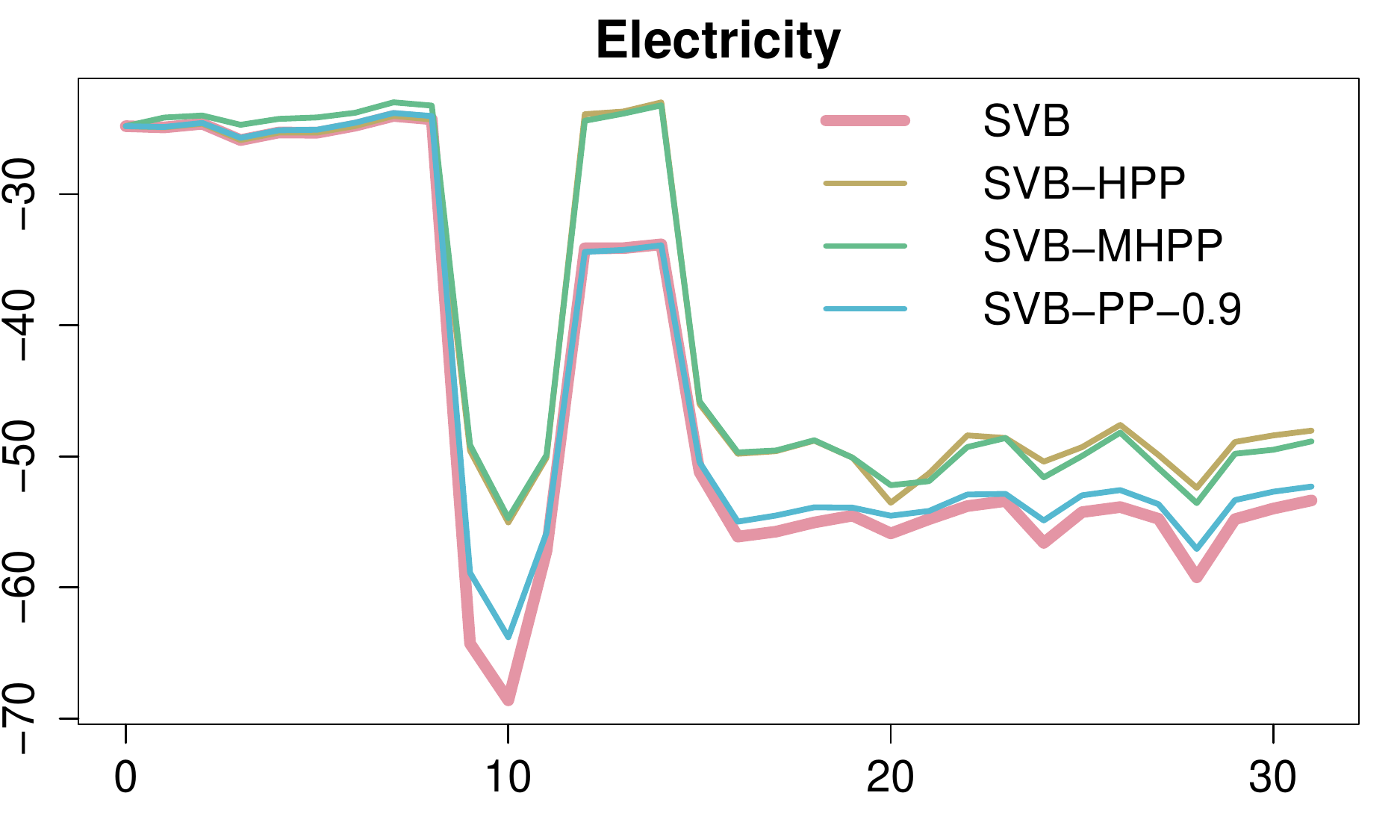}
		\caption{Electricity}
		\label{app:fig:exp:elec}
	\end{center}
	\vskip -0.2in
\end{figure}

\begin{figure}[htb!]
	\begin{center}
		\includegraphics[width=8cm,height=4.5cm,keepaspectratio]{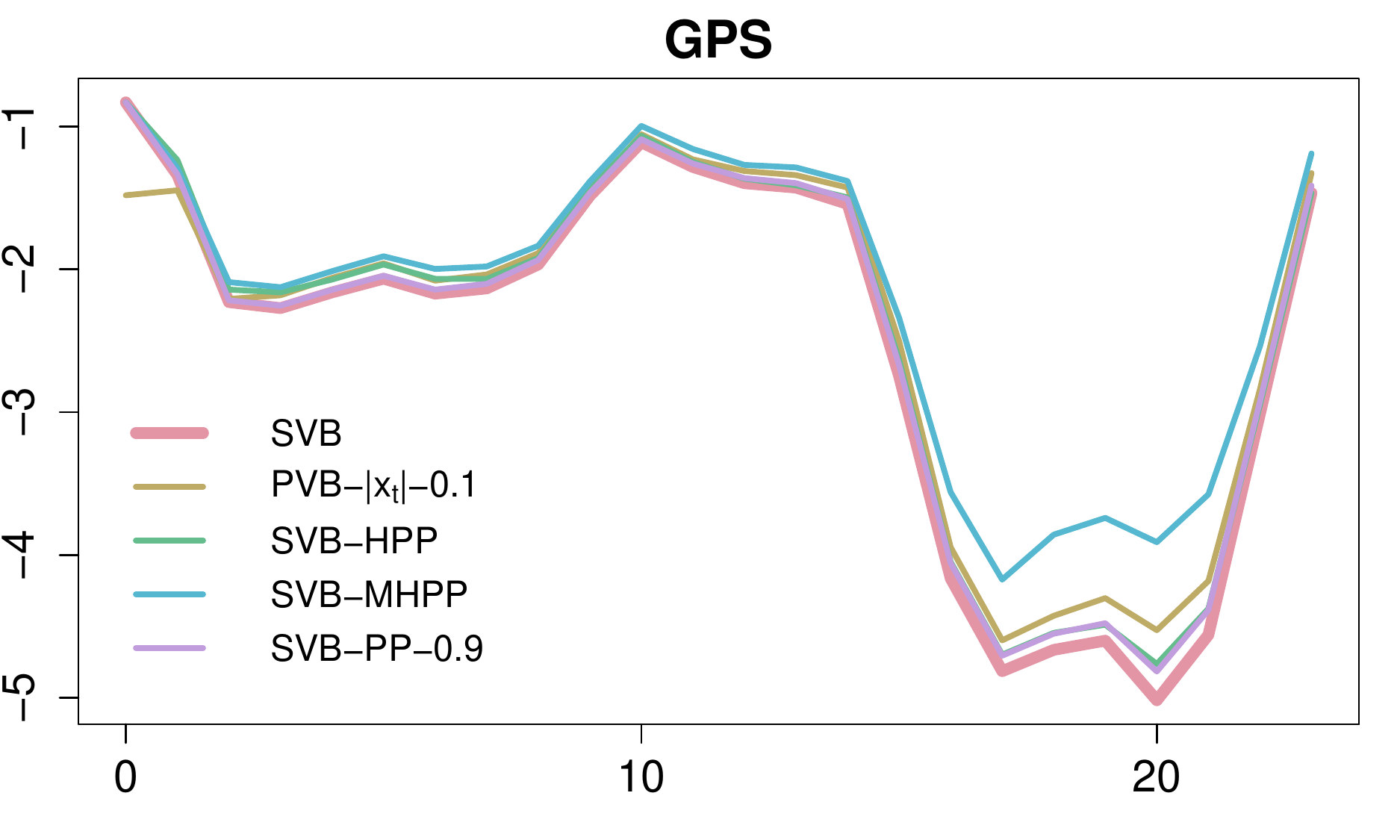}
		\caption{GPS}
		\label{app:fig:exp:gps}
	\end{center}
	\vskip -0.2in
\end{figure} 

\begin{figure}[htb!]
	\begin{center}
		\includegraphics[width=8cm,height=4.5cm,keepaspectratio]{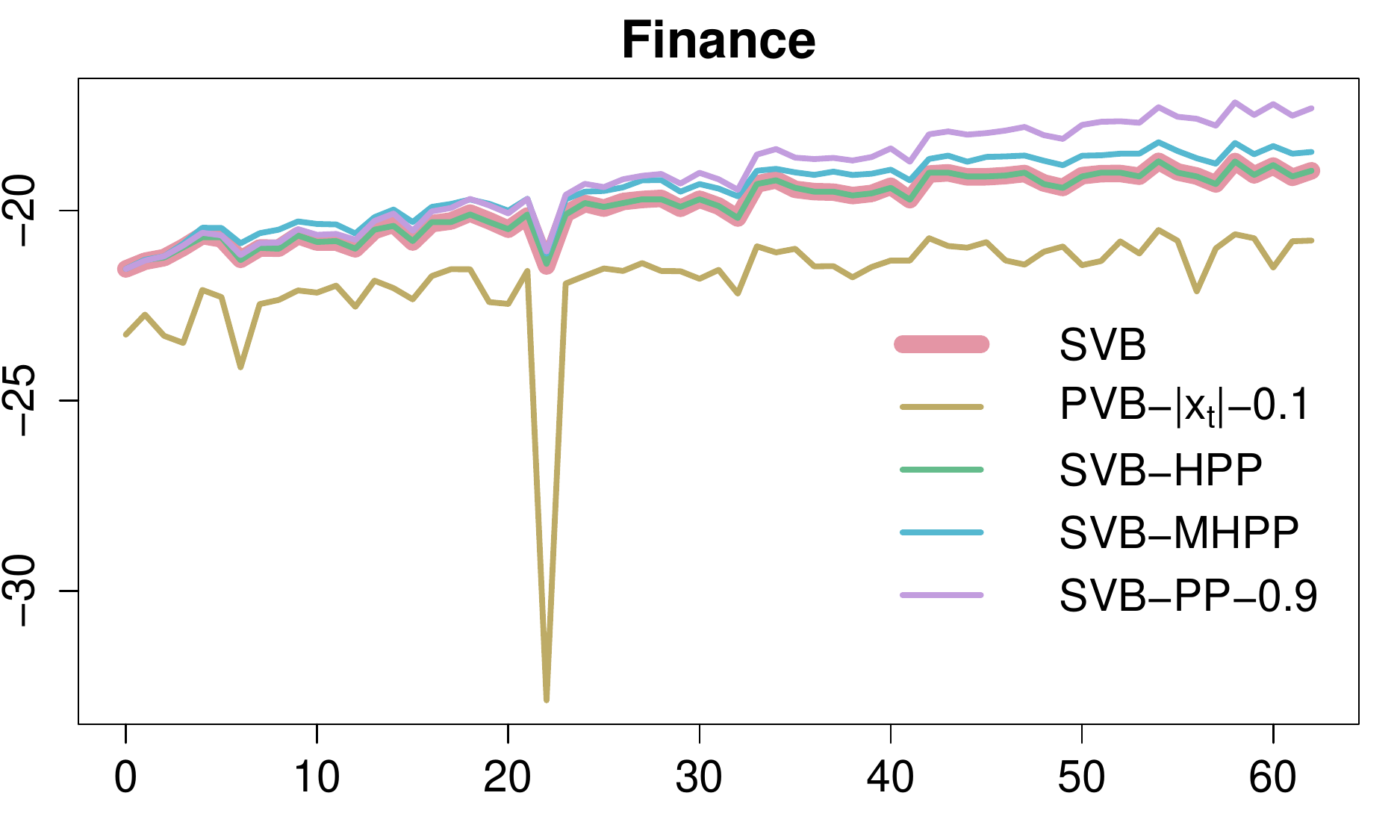}
		\caption{Finance}
		\label{app:fig:exp:bcc}
	\end{center}
	\vskip -0.2in
\end{figure}